\documentclass{ametsocV6.1}[draft]

\usepackage{amsmath,amsfonts,amssymb,bm}
\usepackage{mathptmx}%{times}
\usepackage{newtxtext}
\usepackage{newtxmath}
\usepackage{relsize}
\usepackage{hyperref}

\usepackage{xcolor}

\newcommand{\hide}[1]{ }

\title{Knowledge-Guided Machine Learning: Illustrating the use of Explainable Boosting Machines to Identify Overshooting Tops in Satellite Imagery.}
\authors{Nathan Mitchell\aff{a,b}\correspondingauthor{Nathan Mitchell, {nathan.mitchell24@alumni.colostate.edu}}, Lander Ver Hoef\aff{c}, Imme Ebert-Uphoff \aff{c,d}, Kristina Moen\aff{e}, Kyle Hilburn\aff{c}, Yoonjin Lee\aff{f}, Emily J. King\aff{e}\\ 
}

\affiliation{
\aff{a}{The Courant Institute of Mathematical Sciences, New York University, New York, New York}\\
\aff{b}{Department of Statistics, Colorado State University, Fort Collins, Colorado}\\
\aff{c}{Cooperative Institute for Research in the Atmosphere, Colorado State University, Fort Collins, Colorado}\\
\aff{d}{Department of Electrical and Computer Engineering, Colorado State University, Fort Collins, Colorado}\\
\aff{e}{Department of Mathematics, Colorado State University, Fort Collins, Colorado}\\
\aff{f}{School of Earth and Environmental Sciences, Seoul National University, Seoul, South Korea}
}

\abstract{
Machine learning (ML) algorithms have emerged in many meteorological applications.  However, these algorithms struggle to extrapolate beyond the data they were trained on, i.e., they may adopt faulty strategies that lead to catastrophic failures.  These failures are difficult to predict due to the opaque nature of ML algorithms. In high-stakes applications, such as severe weather forecasting, is is crucial to avoid such failures. One approach to address this issue is to develop more interpretable ML algorithms. The primary goal of this work is to illustrate the use of a specific interpretable ML algorithm that has not yet found much use in meteorology, Explainable Boosting Machines (EBMs). We demonstrate that EBMs are particularly suitable to implement human-guided strategies in an ML algorithm. 
\\
\indent \indent
As guiding example, we show how to develop an EBM to detect overshooting tops (OTs) in satellite imagery. EBMs require input features to be scalar.  We use techniques from Knowledge-Guided Machine Learning to first extract scalar features from meteorological imagery.  
For the application of identifying OTs this includes extracting cloud texture from satellite imagery using Gray-Level Co-occurrence Matrices. 
\\
\indent \indent
Once trained, the EBM was examined and minimally altered to more closely match strategies used by domain scientists to identify OTs. The result of our efforts is a fully interpretable ML algorithm developed in a human-machine collaboration that uses human-guided strategies. While the final model does not reach the accuracy of more complex approaches, it performs reasonably well and we hope paves the way for  building more interpretable ML algorithms for this and other meteorological applications.
}

\begin{document}
\nolinenumbers
%% Necessary!
\maketitle

%%%%%%%%%%%%%%%%%%%%%%%%%%%%%%%%%%%%%%%%%%%%%%%%%%%%%%%%%%%%%%%%%%%%%
% SIGNIFICANCE STATEMENT/CAPSULE SUMMARY
%%%%%%%%%%%%%%%%%%%%%%%%%%%%%%%%%%%%%%%%%%%%%%%%%%%%%%%%%%%%%%%%%%%%%
%
% If you are including an optional significance statement for a journal article or a required capsule summary for BAMS 
% (see www.ametsoc.org/ams/index.cfm/publications/authors/journal-and-bams-authors/formatting-and-manuscript-components for details), 
% please apply the necessary command as shown below:
%
% Significance Statement (all journals except BAMS)
%
\statement{
The purpose of this work is to introduce the interpretable machine learning method of Explainable Boosting Machines (EBMs) to an atmospheric science audience by closely examining how they can be used to detect the location of overshooting cloud tops, which have been associated with the occurrence of severe weather, in satellite imagery. Interpretable machine learning methods are important in high-risk situations such as overshooting top detection as they allow forecasters to have a better understanding of how exactly an identification is made. We walk through how to build, interpret, and modify the machine learning algorithm for use on this task, and discuss other applications that could benefit from this approach. 
%but the steps discussed can be generalized to any application.
}
%
%% Capsule (BAMS only)
%%
%\capsule
%       Enter BAMS capsule here, no more than 30 words. See \url{www.ametsoc.org/index.cfm/ams/publications/author-information/formatting-and-manuscript-components/#capsule} for details.
%
%% * * If using twocol mode, you will need to use the commands "twocolsig" and "twocolcapsule" in place of "sig" and "capsule"
%%      to ensure that the text box correctly spans across both columns.
%

%%%%%%%%%%%%%%%%%%%%%%%%%%%%%%%%%%%%%%%%%%%%%%%%%%%%%%%%%%%%%%%%%%%%%
% MAIN BODY OF PAPER
%%%%%%%%%%%%%%%%%%%%%%%%%%%%%%%%%%%%%%%%%%%%%%%%%%%%%%%%%%%%%%%%%%%%%

\section{Introduction}
\label{sec:intro}
Data-driven algorithms, especially machine learning (ML) algorithms, are competing with physics-based algorithms in many meteorological applications. ML algorithms can bring key advantages: they are often faster, sometimes by orders-of-magnitude \citep{pathak2022fourcastnet,bi2023accurate}, and may score higher on accuracy metrics, e.g., by learning subtle details that may not have been accounted for in a physics-based model directly from vast amounts of observed data.

ML algorithms, however, also bring new risks: they tend to be less transparent than physics-based algorithms (i.e., {\it how} they make decisions is unknown) and, in contrast to physics-based algorithms, they often struggle to generalize beyond the data considered during their development. As a result, ML algorithms may fail catastrophically when applied to data not represented in training, e.g., for extreme or otherwise unusual events. An ML algorithm's {\it failure modes}--the input scenarios under which they fail--are often impossible to predict due to their lack of transparency.
Addressing such failure modes is crucial when dealing with ML algorithms designed for high-stakes applications (e.g., severe weather forecasting, where forecasts are linked to actionable decisions that can affect many lives and properties \citep{NOAA_SP_2022}). Furthermore, lack of trust in ML models is a primary obstacle for their transition into operational use in weather forecasting \citep{boukabara2021outlook,bostrom2024trust}.

Many approaches have been developed to deal with the issue of failure modes. For example, one can attempt to identify these modes using Explainable AI \citep{lapuschkin2019unmasking} or ML uncertainty quantification \citep{psaros2023uncertainty}. (The term AI--or Artificial Intelligence--is used here interchangeably with ML.) Alternatively, one can seek to develop inherently transparent ML algorithms that have minimal unknown failure modes using Interpretable AI \citep{rudin2019stop}, for example by incorporating physics (physics-informed machine learning; \citet{kashinath2021physics,meng2025physics}) or other forms of expert knowledge (knowledge-guided machine learning (KGML); \citet{karpatne2022knowledge}) into ML algorithms. Our team is part of the NSF AI Institute for Research on Trustworthy AI in Weather, Climate, and Coastal Oceanography (AI2ES)  \citep{mcgovern2022nsf,mcgovern2024ai2es}, which is exploring many of those approaches. Here, we focus on the use of Interpretable AI in the form of Explainable Boosting Machines (EBMs), combined with KGML.
\subsection{Failure Modes and Explainable AI}
\label{sec:Clever_Hans_sec}
Failure modes of ML algorithms arise from many different sources, including limitations in the chosen modeling framework (explored in Section \ref{sec:results}) and from spurious correlations in the training data (i.e., correlations that exist in the data set but are not representative of the system the data are meant to represent--explored in this section). For a comprehensive list of potential issues along with real-world examples in environmental science, see \citet{mcgovern2022we}.

\citet{calude2017deluge} showed that {\it all data sets} contain spurious correlations. Fully data-driven ML methods are prone to developing faulty strategies based on such spurious correlations. Those strategies are often referred to as {\it Clever Hans} strategies, based on the seminal work by \citet{lapuschkin2019unmasking}, and can lead to failures in real-world use. Thus, any ML model is susceptible to adopting faulty strategies. The more complex the model, the harder it is to identify the resulting failure modes \citep{rudin2019stop}. 

The field of {\it Explainable AI (XAI)} was created to retrospectively identify an ML model's learned strategies using post-hoc explanations on otherwise black-box models.
\begin{figure}
    \centering
    \includegraphics[width=0.9\linewidth]{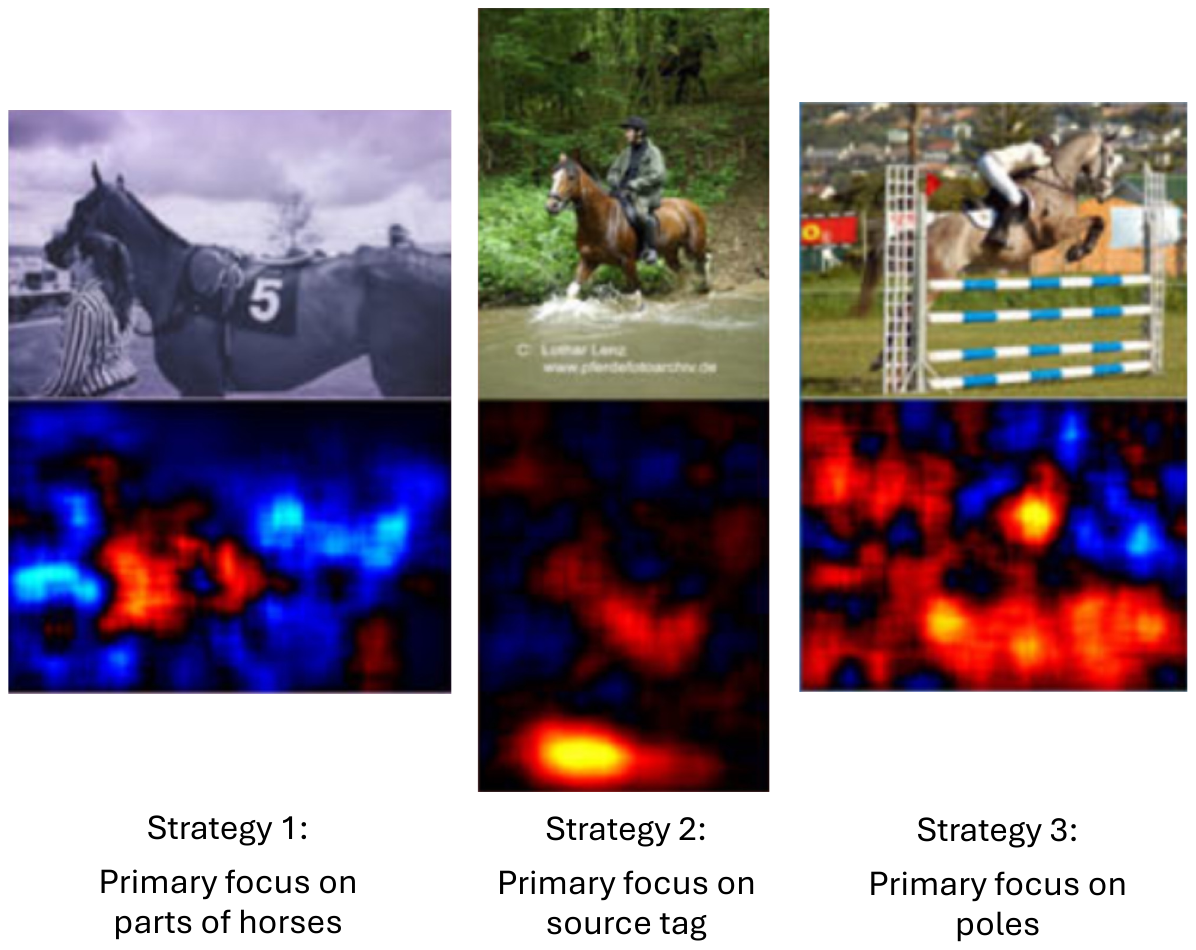}
    \caption{Illustration of Clever Hans Strategies by  \citet{lapuschkin2019unmasking}. 
    These examples illustrate three strategies of an AI algorithm tasked to decide whether a given image contains a horse.  
    Input images of the algorithm are shown on top. The 
    corresponding attribution maps generated by Layer-Wise Relevance Propagation are shown on the bottom.  These maps are overlays indicating which pixels in the original image contribute positively (red) or negatively (blue) to the AI system deciding that the image contains a horse. 
    Strategy 1 is appropriate, while Strategies 2 and 3 are faulty.\\  
    Image credit: Adapted from Fig.\ 3 in \citep{lapuschkin2019unmasking}.}
    \label{fig:Clever_Hans}
\end{figure}
Figure \ref{fig:Clever_Hans}--which is based on Fig.\ 3 in \citet{lapuschkin2019unmasking}--illustrates the use of the XAI method of Layer-Wise Relevance Propagation (LRP; \cite{bach2015pixel}) to identify failure modes of a neural network algorithm trained to identify the presence of objects, such as horses, within an image. Fig.\ \ref{fig:Clever_Hans} shows three strategies learned by the neural network to identify horses: one where the algorithm focuses on actual attributes of a horse (a correct strategy), one where the algorithm focuses on a source tag left on a subset of the horse images (a faulty strategy), and one where the algorithm focuses on poles within the image (another faulty strategy). These strategies are revealed by use of LRP attribution maps that estimate which areas in the input image contribute positively (red) or negatively (blue) to the algorithm claiming that a horse is present. In a real-world application, the second and third strategies would, respectively, result in misses (not predicting a horse when the tag is not present) and false alarms (predicting a horse when poles are present but a horse is not). Here, the XAI method successfully identified two strategies that would have led to failures if the model was deployed as-is.

XAI methods have also been utilized in many atmospheric applications; see, e.g., \citet{mcgovern2019making,toms2020physically,flora2024Explainability,bommer2024finding,fan2024physically,krell2024using}, and references therein. Following the approach by \citet{lapuschkin2019unmasking}, \citet{GREMLIN} used the LRP method to identify the strategies of an algorithm developed called GREMLIN--a convolutional neural network (CNN) used to estimate Multi-Radar Multi-Sensor (MRMS) composite reflectivity \citep{Smith2016MRMS} from Geostationary Operational Environmental Satellite (GOES) imagery \citep{Schmit2017CloserLook,Goodman2013GOES}. Using LRP allowed for the identification of four primary strategies that GREMLIN uses to estimate high values of radar reflectivity. All of them appeared meaningful.

XAI methods, however, are limited in their ability to identify failure modes. For example, attribution maps, such as LRP, are limited to identifying strategies that a model employed for a \emph{specific} sample and typically require human interpretation of the attribution map of each sample. This makes establishing a broad understanding of all strategies and failure modes across a complex model extremely challenging. \citet{mamalakis2022investigating,mamalakis2023carefully} discusses additional issues for attribution methods, while \citet{rudin2019stop} discusses general short comings of XAI methods. 
\subsection{Interpretable AI and Knowledge-Guided Machine Learning}
\label{sec:IAI_intro}
Another way to identify failure modes is {\it Interpretable AI} which seeks to build understanding into an ML model from the start, rendering the use of post-hoc explanation methods (XAI) unnecessary \citep{flora2024Explainability,rudin2019stop}. In contrast to Explainable AI, Interpretable AI methods yield increased transparency of the ML model's strategies and thus a much deeper and more complete understanding of the model's failure modes. 

Unfortunately, building an interpretable model is not possible for all ML tasks, as some tasks are too complex to allow for an interpretable model. The ML component of many tasks can be simplified, however, by incorporating prior knowledge, an approach known as {\it knowledge-guided machine learning (KGML)} \citep{karpatne2022knowledge,karpatne2024knowledge}. KGML is the approach taken here, seeking to combine the best of both worlds--human expertise and data-driven algorithms--to yield algorithms that implement human-approved strategies while also learning subtle details directly from data. Users of this approach should be aware that the increased transparency of Interpretable AI methods using KGML comes with trade-offs compared to standard black-box ML models:  

\begin{enumerate}

\item {\bf Upfront vs. Backend Time Commitment:}\\
    Building Interpretable AI methods with KGML requires developing a custom approach for {\it each} application, i.e., considerable {\it upfront} time and effort investments.\\
    In contrast, black-box models skip these {\it upfront} time commitment, but often have higher {\it backend} time commitments, e.g., to apply XAI after development to gain useful insights into the model--often with limited success as discussed in Section \ref{sec:intro}\ref{sec:Clever_Hans_sec}.
    
\item {\bf Reduced Accuracy vs. Fewer Catastrophic Errors:}\\
    Interpretable ML models enforce model simplicity that may, in some instances, prevent the model from capturing all details in the data and thus lead to reduced \emph{average} accuracy when compared to more traditional deep learning models.
    Developers need to carefully weigh for their specific application the trade-off between potentially reduced average accuracy against reduced risk of {\it catastrophic} errors.
\end{enumerate}
\subsection{Explainable Boosting Machines}
We explore a type of Interpretable AI method known as Explainable Boosting Machines (EBMs) which were introduced by \citet{Nori2019InterpretMLAU}. EBMs--in spite of the term ``Explainable'' in their name--are simple and \emph{interpretable} models. As we will explore later, because of their modular nature, EBMs make it easy to directly edit--and thus correct--any strategies identified to be faulty. We refer to this subset of Interpretable AI algorithms as {\it Editable AI} and discuss it more in Section \ref{sec:introducing_EBMs}.

EBMs have brought many advancements to medical, health care, and financial applications (see Section \ref{sec:introducing_EBMs}\ref{sec:EBM_literature}). This paper evaluates the use of EBMs, in combination with feature engineering, to obtain an interpretable, knowledge-guided ML algorithm for analyzing meteorological satellite imagery. Our approach was inspired by a talk given by \citet{Caruana2023} on the benefits of EBMs. Motivated by this talk, we set out to explore whether EBMs can also be applied to meteorological applications. 
We emphasize that the goal of this study is not to deliver a new, state-of-the-art algorithm for this application capable of deployment. Instead, our primary goal is to explore and illustrate EBMs using this application as guiding example, and thus to make the meteorological community aware of EBMs' unique capabilities and encourage further exploration. 
\subsection{Organization of this article}
Because this article's goal is the popularization of EBMs using a case study, it has a slightly non-standard organization--one outlined here:

\begin{itemize}
    \item {\bf Section \ref{sec:intro}} %we have  
    provides an introduction to interpretable AI, noting why it is important and that EBMs are a tool in that space.
    \item {\bf Section \ref{sec:introducing_EBMs}:} provides a more thorough account of EBMs, detailing what an EBM is, where they have been used in the past, and where we see them being used in the future.
    \item{\bf Section \ref{sec:overshooting_tops_intro}:} introduces the chosen application, detecting the locations of OTs in satellite imagery, and describes the dataset used and the feature engineering performed.
    \item {\bf Section \ref{sec:methods}:} discusses the practical usage of EBMs via visualization and interpretation followed by a discussion of model editing.
    \item {\bf Section \ref{sec:results}:} discusses model performance via overall results and representative samples.
    \item {\bf Section \ref{sec:conclusions}:} presents a final discussion and our conclusions.
\end{itemize}

\section{Introducing Explainable Boosting Machines}
\label{sec:introducing_EBMs}
EBMs are simple models that can be thought of as an extension of the multiple linear regression scheme. Linear regression can be expressed as: 
\begin{equation}
    Y = \beta_0 +{\sum_i} \beta_i X_i + \epsilon,
    \label{eq_lin_regression}
\end{equation}
where the response $Y$ is governed by an intercept $\beta_0$; slopes $\beta_i$'s, each corresponding to an independent predictor $X_i$; and an error term $\epsilon$. For each $i$, $\beta_i X_i$ is a linear function representing the contribution of predictor $X_i$. Thus, the linear regression model can be viewed as a sum of linear functions of the predictors. Similarly, EBMs can be viewed as a sum of \emph{non-linear} functions. Instead of a slope, each feature is associated with a non-linear function (known as a {\it feature function}) that describes the relationship between the feature and the outcome of interest.

When only univariate (1D) functions are considered--which we refer to as the ``primary'' feature functions--EBMs reduce down to Generalized Additive Models (GAMs, \cite{Hastie1987Generalized}), which is not a coincidence; EBMs were built upon the GAM framework to address the gap in performance between GAMs and more complex models \citep{Lou2013AccurateIntelligible,Lou2012Intelligible}. The bolstered performance comes from the inclusion of bivariate (2D) feature functions--which we refer to as the ``interaction'' feature functions--which are not possible under the GAM framework. Even with interactions, EBMs remain highly intelligible.

In general, EBMs take the form:
\begin{equation}
     g(E[y]) = \beta_0+\mathlarger{\mathlarger{\sum_i}}f_i(x_i)+\mathlarger{\mathlarger{\sum_{i, j \in I}}}f_{i_j}(x_i,x_j),
     \label{eq_EBM}
\end{equation}
where $f_i(x_i)$ denotes a primary feature function representing the contribution of feature $x_i$ to the outcome and $f_{i_j}(x_i,x_j)$ denotes an interaction feature function representing. The GA$^2$M algorithm, of which EBM is an implementation \citep{Nori2019InterpretMLAU}, learns the main effect feature functions first using bagging and gradient boosted trees \citep{Caruana2015IntelligibleHeatlCare}. During the boosting process, only one feature function is trained at a time in a ``round-robin'' fashion to appropriately deal with features that are potentially linearly dependent and to ensure feature functions that show the appropriate independent, modular relationship with the response are trained \citep{Nori2019InterpretMLAU}. Additionally, a low learning rate is used to ensure that the order in which the features are included and thus the order in which the functions are learned has minimal impact on the learned model \citep{Nori2019InterpretMLAU}.

After the main effect feature functions have been trained, the algorithm detects then ranks pairwise interactions in the residuals. Cross-validation is used to select the (limited) number of pairwise interactions to be included in the model \citep{Caruana2015IntelligibleHeatlCare}. More information on this algorithm can be found in \cite{Lou2013AccurateIntelligible}. Here, we represent this set with $I$. Though it is possible to manually include higher-order interactions within the model, they are not automatically included. Finally, \textit{g} represents the chosen link function which allows for tasks such as regression or classification to be performed \citep{Nori2019InterpretMLAU}, and ${\beta}_0$ represents the model's learned intercept (a single, scalar value).

These non-linear feature functions are designed to be strongly interpretable. A one-dimensional function is a structure that is simple enough to be easily plotted and understood by a user, while being complex enough to collect many simpler terms (such as linear or quadratic regression terms, or decision tree branches) into a small enough number of feature functions that the overall strategy of the model can be interpreted.

In this work, we exploit the editable nature of EBMs. To explain how we do this, it is worth diving a little deeper on what the feature functions are. Each function is a step function that acts like a lookup table; for a given range of input values (the {\it key}), a single output value is returned (the {\it value}). For every feature function, the keys are determined from the data and the values are learned.

To classify any given pixel (which corresponds to a set of input scalar values), a simple process is followed: for every feature function, pull the value from the key the pixel fits into. Then, add the values (also called ``scores''; units are log-odds) together along with the intercept. Once a final score has been calculated, the link function (inverse logit) is applied to generate a probability. If greater than or equal to 0.50, the pixel is assigned the ``event'' class. If less than 0.50, the pixel is assigned the ``non-event'' class.

Altering an EBM, then, is merely changing the learned score values (i.e., requiring the keys to map to a different, user-defined value) of a given feature function. Furthermore, the intercept can be changed. This alteration does not impact the model's strategies. Instead, raising the intercept is equivalent to lowering the threshold at which a prediction is made and vice versa for lowering the intercept. Ultimately, EBM alteration amounts to changing which regions in the feature space should or should not result in a detection, a form of granular ``local'' strategy editing that is not possible even with simpler models like linear regression.

\subsection{How EBMs Enable Knowledge-Guided Machine Learning}
\label{sec:ebms_enable_learning}
\begin{figure*}%[htp]
\centering
\noindent\includegraphics[width=1\textwidth,angle=0]{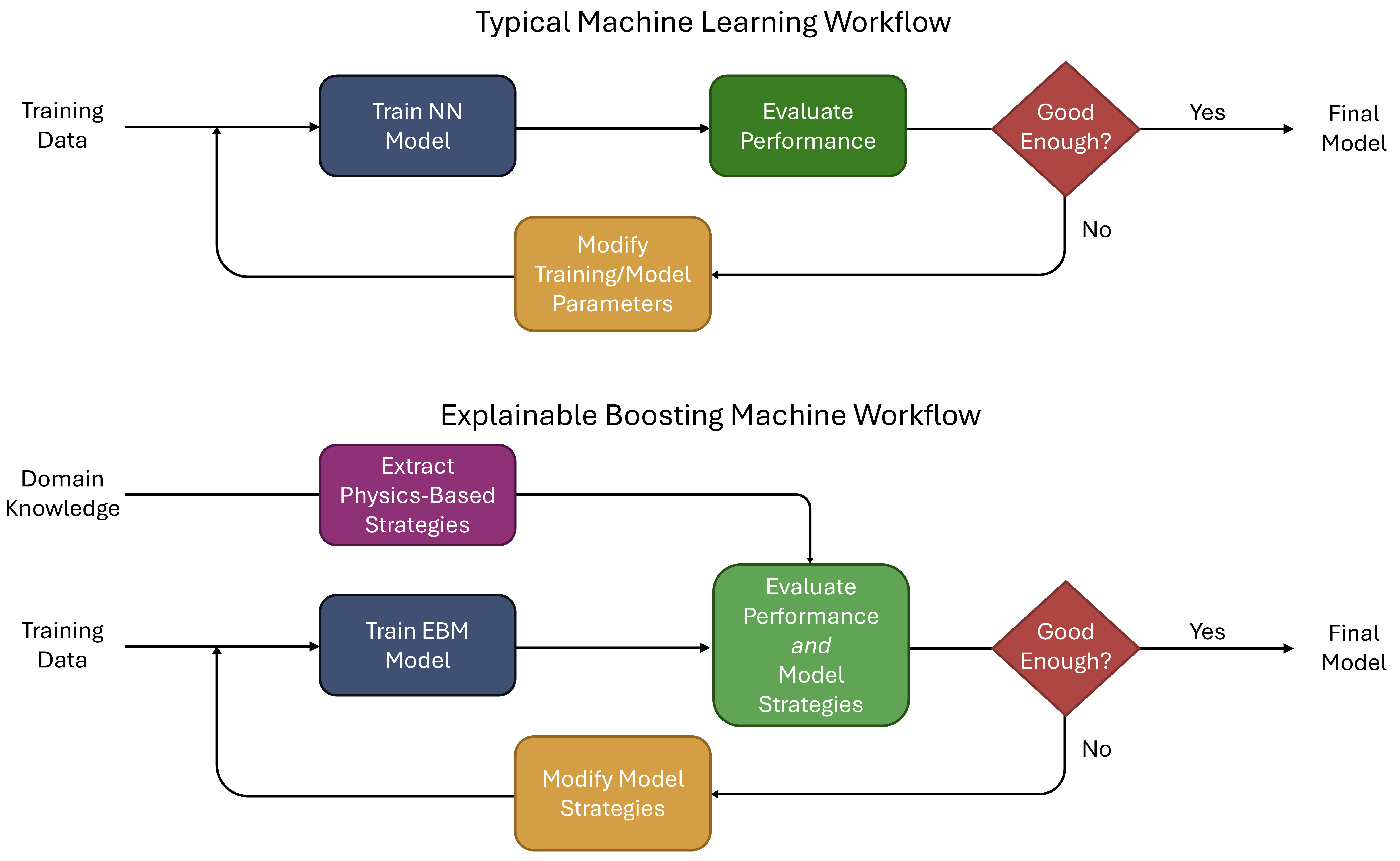}
\caption{Examples of two ML workflows--(a) for black-box ML models, such as neural networks, and (b) for EBMs. Note that in (b) the model's strategies are modified without re-training the model.\label{workflows}}
\end{figure*}

Given the novelty of this approach, we illustrate how we envision implementing the interpretability and editability of EBMs into an ML workflow. Fig.\ \ref{workflows} compares two workflows: one for conventional ML methods--such as a CNN or random forest--in Fig.\ \ref{workflows}a, and one for EBMs in Fig.\ \ref{workflows}b.
As Fig.\ \ref{workflows}a illustrates, a common loop for conventional ML applications involves evaluating and retraining models. \cite{Caruana2015IntelligibleHeatlCare} refer to this behavior as ``repairing'' a model. Hyperparameters may be adjusted or the data may be altered, but, in order to \emph{change} the model, the model (or parts of it) must be retrained. In the case of black-box models, there is no way to guarantee changes lead to a desired effect nor that these changes made the underlying strategies more physically realistic. XAI and some Interpretable AI methods share the first shortcoming and are an attempt to at least partially address the second.

EBMs are not susceptible to either, giving rise to a new loop--one without model retraining--illustrated in Fig.\ \ref{workflows}b. An EBM's strategies can be compared to known strategies and adjusted accordingly, all without retraining. Such adjustments can be made until a desired model has been constructed. We incorporate this into the third step of our three-step plan for constructing a KGML approach:
\begin{itemize}
\item[] 
    {\bf Step 1:} Identify and Incorporate Expert Knowledge\\
    Identify strategies an expert would use to solve the task then engineer features that encourage the EBM to follow them.
\item[]
    {\bf Step 2:} Interpretable AI\\ 
    Train an EBM using training data that uses the features from Step 1. 
\item[]
    {\bf Step 3:} Editing the strategies\\
    Test performance, and visualize and adjust strategies as needed.
\end{itemize}
\subsection{Prior Use of EBMs and Proposed New Meteorological Applications}
\label{sec:EBM_literature}
EBMs have found frequent use in the health sciences (see \citet{morgan2021explainable,sarica2021explainable,hegselmann2022development,wang2022interpretable,patel2023explainable,korner2024explainable,arslan2024combining,wang2024interpretable,yagin2025explainable} and many other studies). We could find only one meteorological application of EBMs where they were used to estimate wind shear for aviation tasks \citep{khattak2023assessing, khattak2023explainable}. These studies focus primarily on EBM application and, as such, do not provide a thorough tutorial on the theory and implementation of EBMs. Moreover, they do not make use of the ability to edit EBMs. Elsewhere in environmental science, EBMs have been used to anticipate landslides \citep{caleca2024shifting}, while in ecology they have been used to estimate the date of leaf unfolding \citep{gao2024interpreting} and in agriculture to predict crop yield \citep{celik2023informative,pant2025comparative} and pests \citep{nanushi2022pest}.

We see a wide range of opportunities in meteorology to make use of EBMs, which we will discuss in brief here. The ultimate suitability of EBMs for a particular application can only be determined (as with all machine learning algorithms) by trial and error, but we hope to promote their inclusion in the set of algorithms worthy of being tried.

In this application we use EBMs on image data, but EBMs were first developed for tabular, non-spatial data. As such, they would be very suitable for application to one of the more well-known tabular datasets in tropical cyclone meteorology, the Statistical Hurricane Intensity Prediction Scheme (SHIPS) dataset \citep{demaria1994ships}. Many studies and algorithms have utilized these data as inputs for predictions of (for instance) tropical cyclone rapid intensification using both statistical methods \cite{rozoff2011new} and more advanced AI methods \citep{lagerquist2025identifying}. EBMs could provide a middle ground between these approaches, providing the interpretability of statistical methods and the power and flexibility of ML methods, while also allowing the many human forecasters avenues to incorporate their knowledge directly into the algorithm.

We also see EBMs being useful for tasks combining information from multiple channels of satellite data (i.e., \emph{multispectral} imagery). Identifying dust plumes (and particularly distinguishing them from clouds), for instance, is one that has most commonly been solved using multispectral imagery (as in \cite{miller2017debra}). 

Image-to-image translation is a task that also relies on multispectral information. For instance, recent work \citep{haynes2025creating} has shown that it is possible to create synthetic passive microwave data from geostationary infrared data. That work tested both CNNs and diffusion models, both of which are black-box. To apply EBMs to this task would require feature engineering to encode spatial information, but we would be interested to see how EBMs fare.

In addition to this small selection, there are many more projects that could benefit from the use of EBMs. We hope this paper will make EBMs more accessible to meteorologists so they are more able to make use of them in their own projects.
\section{Overshooting Top Detection--Introduction to the Problem and a Proposed Solution using EBMs}
\label{sec:overshooting_tops_intro}
As a first sample application, we explore how EBMs can be used to detect the location of overshooting tops (OTs) in satellite imagery. This task has been tackled multiple times before \citep{Bedka2010Objective, Bedka2016Probabalistic, Khlopenkov2021Recent}, most recently by using traditional black-box deep learning models \citep{Kim2017Detection, Cooney2025AutomatedDetection}. 

More specifically, satellite imagery has been used to connect OT signatures with the occurrence of severe weather events such as heavy rainfall, strong winds, hail, and tornadoes \citep{negri1981Rainfaill, Heymsfield1991SevereStorms, Brunner2007SevereWeather, Dworak2012SevereStormReports, Mikus2013OTWeatherConditions}. Severe storms--those producing hail, tornadoes, and straight-line winds--are an increasing contributor to billion-dollar disasters in the United States. In their original analysis, \cite{billion_dollar_disasters_2013} found that, of the billion-dollar disasters between 1980 and 2011, severe storms made up 32\% of the events. Additional analysis by \cite{billion_dollar_disasters_website} has shown, from 1980 to 2023, severe storms now make up nearly 50\% of the billion-dollar disasters with a strong increase in the severe storm count starting in 2006. 
\begin{figure*}%[htp]
\centering
\noindent\includegraphics[width=1\textwidth,angle=0]{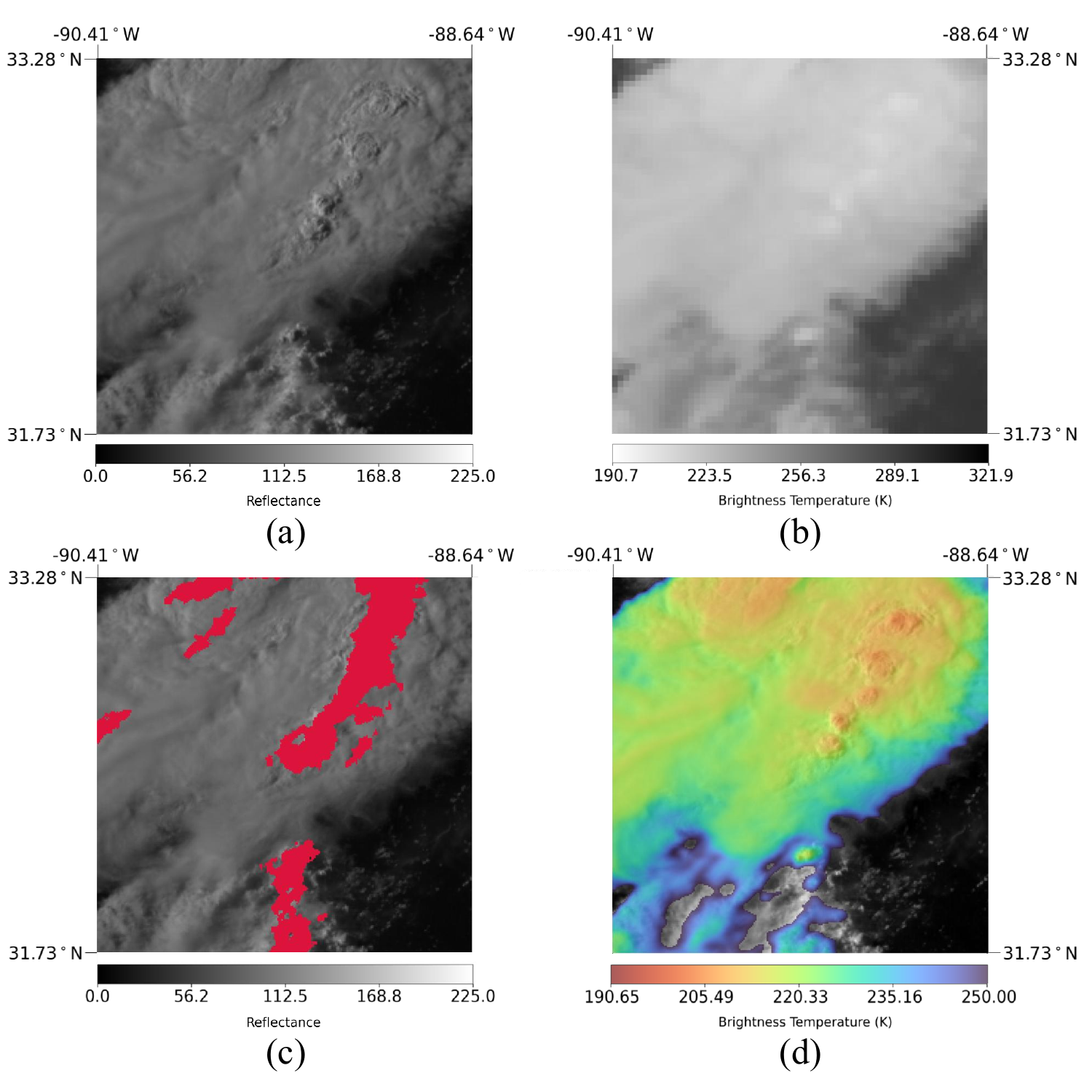}
\caption{Examples of (a) visible channel reflectance, (b) infrared channel brightness temperature, (c) MRMS labels (in red), and (d) IR/VIS sandwich product, each from 5 June 2024 at 21:45:00Z.\label{VisibleInfraredMRMS}}
\end{figure*}

Given radar coverage limitations (see Fig.\ 1 of \citep{lee2024validating}), satellite imagery provides a way to identify OTs globally. The latest generation of geostationary satellite imagers are ideally suited to this task. In particular, we utilize data from the GOES-16 satellite. The Advanced Baseline Imager (ABI) \citep{ABI_BAMS_paper} aboard GOES-16 has a spatial resolution ranging from 0.5 to 2 km and a temporal refresh rate ranging from 0.5 to 15 minutes. These scales are capable of observing features associated with OTs, as illustrated by recent work by \cite{Cooney2025AutomatedDetection}.

Multi-spectral information enhances situational awareness. We consider reflectance (shown in Fig.\ \ref{VisibleInfraredMRMS}a) derived from visible (VIS) imagery as well as infrared (IR) imagery (shown in Fig.\ \ref{VisibleInfraredMRMS}b). Reflectance values show the cloud-top bubbling associated with OTs while brightness temperature values from the IR imagery can be used as a proxy for cloud-top height and can show where OTs are located.

Information from these two channels can be combined by overlaying partially transparent, color-enhanced IR imagery at temperatures less than or equal to 250 K over VIS imagery to create a IR/VIS sandwich product, shown in Fig.\ \ref{VisibleInfraredMRMS}d. Shown in Fig.\ \ref{VisibleInfraredMRMS}c, the “PrecipFlag” product from the Multi-Radar Multi-Sensor (MRMS) system can be used to provide insight into the location of convection.
\subsection{Existing Approaches to Identify Overshooting Tops}
\label{sec:existing_approaches}
Numerical thresholds are a classic scheme for isolating and identifying OTs. When using satellite data, thresholds derived from IR imagery are most common as that imagery does not rely on solar illumination \citep{Kim2017Detection}. Fixed brightness temperature \citep{Ai2017DeepConvective} and infrared window channel texture \citep{Bedka2010Objective} values, among others, can be used. The value, however, depends on many factors (e.g., the satellite, intensity of the convective updraft) \citep{Bedka2010Objective}, meaning there is not a one-size-fits-all value that can be chosen to reliably separate OTs from their surroundings in all cases.

Other approaches to OT identification have utilized ML techniques such as CNNs, including work done by \cite{Kim2018DeepLearning}. CNNs provide a natural choice for extracting information from satellite imagery because of their ability to learn complex statistical relationships and make use of the information content in gradients and, more generally, spatio-temporal patterns. More specialized CNNs designed for semantic segmentation, often using a U-net architecture, have been used in this domain. For example, \cite{Cooney2025AutomatedDetection} used the original U-net \citep{Ronneberger2015U-Net}, MultiResUnet \citep{Ibtehaz2020MultiResUNet}, and AttentionUnet \citep{Oktay2018AttentionUnet} architectures to identify OTs within satellite imagery. While effective, these models are black-box, and thus face the discussed limitations.
\subsection{Proposed Strategies to Identify Overshooting Tops}
\label{sec:strategies_section}
Before building an interpretable ML model to automatically detect OTs from satellite-based imagery, we first consider how a human analyst, when presented with satellite imagery, might identify OTs. The product of strong atmospheric instability and subsequent vertical updrafts, OTs can be visually identified by considering the relationship they have with their surroundings. We present two primary strategies to identify OTs:
\begin{itemize}
\item 
   Strategy 1 seeks to identify OTs by considering the elevation difference seen between them and the surrounding cirrus anvil clouds by numerically quantifying this displacement using a proxy such as cloud-top temperature. 
\item
   Strategy 2 involves numerically quantifying texture within regions (typically represented by small groups of pixels) of visible imagery to then isolate the bumpy, textured OTs from the surrounding flatter, less-textured anvil cloud. 
\end{itemize}
We combine these two strategies by considering both cloud-top temperature derived from infrared imagery and texture derived from visible imagery. Additionally, we consider a measure of cloud-top brightness. Conventional knowledge gained from the repeated viewing of satellite imagery tells us, with ample daylight (and especially in the local afternoon), OTs are typically bright and visually stick out from the surrounding anvil in VIS imagery. The brightness channel can also indicate the presence of a shadow cast by an OT, another easily identifiable feature of the phenomenon \citep{Bedka2010Objective}.
\subsection{Steps to Build an Interpretable Model for OT Detection}
\label{sec:toward_interpretable_model}
\begin{figure*}%[htp]
\centering
\noindent\includegraphics[width=1\textwidth,angle=0]{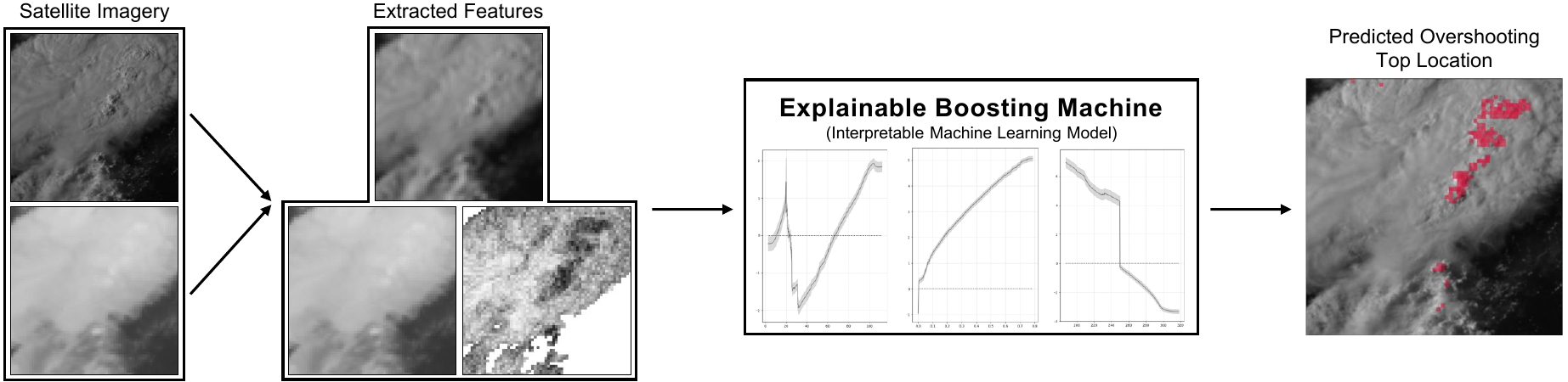}
\\
(a) \hspace*{3.20cm} (b) \hspace*{4.65cm} (c) \hspace*{4.225cm} (d)
\caption{ Summarized approach displaying (a) the satellite imagery used to generate the three features used in our approach (taken 5 June 2024 at 21:45:00Z), (b) the three features themselves (first row: brightness, second row: infrared followed by cool contrast tiles), (c) the interpretable machine learning method used (EBM) and a visualization of three of its learned strategies, and (d) a map of OT locations (in red) as detected by the finalized model overlaid on the corresponding visible imagery. 
\label{overall_workflow}}
\end{figure*}

We propose combining the steps outlined below--and illustrated in Fig.\ \ref{overall_workflow}--to achieve an OT identification algorithm that is both simpler and fully interpretable:
\begin{itemize}
    \item[] {\bf Step 1a:} Use GOES-16 ABI visible and infrared imagery (Fig.\ \ref{overall_workflow}a),
    \item[] {\bf Step 1b:} Derive information-rich features using (a) standard image processing techniques and (b) the mathematical framework of Gray-level Co-occurrence Matrices (GLCMs) to extract texture information using the approach by \citet{Moen2024Gray} (Fig.\ \ref{overall_workflow}b and Section \ref{sec:overshooting_tops_intro}\ref{sec:input_features}),
    \item[] {\bf Step 2:} Combine extracted features using the interpretable ML algorithm Explainable Boosting Machines (Fig.\ \ref{overall_workflow}c) to identify the location of OTs (Fig.\ \ref{overall_workflow}d).
\end{itemize}
Step 1(a,b) and Step 2 above implement for this specific application Step 1 and Step 2, respectively, of the general three-step KGML approach proposed in Section \ref{sec:introducing_EBMs}\ref{sec:ebms_enable_learning}.
\subsection{Data and Feature Extraction}
\label{sec:data}
The dataset used in this application was created in-house and was generated to be similar to the dataset used by \citet{Lee2021Applying}. For our approach, it was separated into (1) a training set---to train the EBM, (2) a validation set---to examine how the model performed and to inform changes to be made to the learned strategies, and (3) a test set---to test the final model on unseen data. These sets were comprised of ``scenes,'' each scene being $64 \times 64$ (4,096) pixels. In total, we had access to 10,404 scenes---5,206 for training, 2,579 for validation, and 2,619 for testing.

Data were collected over the central and eastern parts of the contiguous United States, more specifically from $70^{\mbox{o}}$ W to $105^{\mbox{o}}$ W and from $30^{\mbox{o}}$ N to $48^{\mbox{o}}$ N. As is common in earth science applications, our training, validation, and test datasets came from different years to avoid the impact of significant autocorrelation between samples in the three sets. The training set spans from May to August of both 2021 and 2022, the validation set from May to August of 2023, and the test set from May to August of 2024. We selected the six-hour window within each day with the largest number of storm reports, resulting in 2,898 hours of data. The training set corresponds to 242 distinct storm outbreaks, the validation set 118, and the test set 123. Given these data were collected only during the northern hemisphere's summer months, a limitation of the proposed algorithm is that it has not seen data from other time periods or locations. Performance may suffer if it encounters such data.

Satellite imagery was obtained from GOES-16's ABI, more specifically channel 2 (0.64 $\mu$m) in its native 0.5 km resolution and channel 13 (10.3 $\mu$m) in its native 2 km resolution. Convective flags from data provided by the MRMS system were used as labels to train the model. How information provided by the MRMS system was used to generate these labels is discussed in Section \ref{sec:overshooting_tops_intro}\ref{sec:mrms}.

Because our OT identifier uses visible imagery, it is more limited in scope than other state-of-the-art satellite-based identifiers as it is only usable during the day. To avoid the terminator region, any images taken when the solar zenith angle was over $65^\circ$ were removed from the dataset following choices made by \cite{Lee2021Applying}. This resulted in 797 scenes being removed from the training set, 360 from the validation set, and 447 from the test set.
\subsection{Developing Information-Rich Input Features for Overshooting Top Identification}
\label{sec:input_features}
EBMs have been designed to provide interpretable ML algorithms for tasks that involve scalar values as inputs and outputs. Thus, to use this approach for our application we had to first extract expressive scalar features from the imagery. In total, we extracted three features from the satellite imagery. While temporal information from image {\it sequences} may be useful for detecting OTs (see \cite{Lee2021Applying}, who used temporal information to improve their convection detection), for this first study we only leveraged spatial features.

Below we define the features we extracted. Though we present these features using ``scenes,'' we stress that the EBM algorithm only sees single-pixel ``scalar'' values.
\subsubsection{Brightness Feature}
The first feature, $x_1$, was derived from ABI channel 2 imagery. Channel 2, also known as the ``red'' band, provides high-resolution visible imagery. Visible imagery provides a measure of the intensity of the light being reflected back to the satellite by the clouds and the surface being imaged. For this application, following choices made by \citet{Lee2021SimplifiedConvection}, raw radiance values were converted into reflectance factor values using the $\kappa$ factor, then normalized using the solar zenith angle to obtain reflectance values. Example reflectance data can be seen in Fig.\ \ref{VisibleInfraredMRMS}a.
\begin{figure*}%[htp]
\centering
\noindent\includegraphics[width=0.50\textwidth,angle=0]{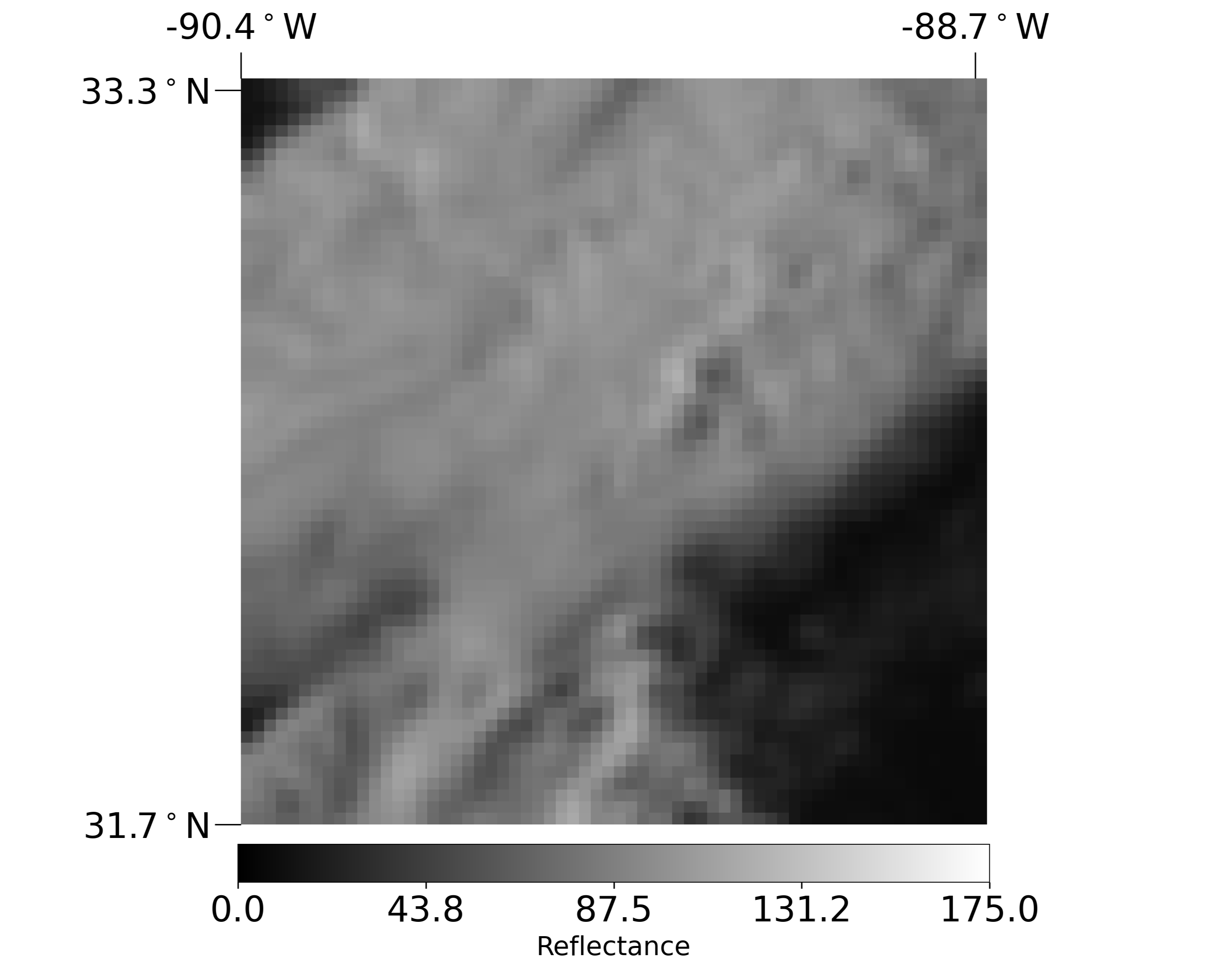}
%\appendcaption{4}
\caption{Brightness feature example. Derived from VIS imagery (for this figure, taken 5 June 2024 at 21:45:00Z). This scene represents the blurred and downsampled version of the VIS imagery displayed in Fig.\ \ref{VisibleInfraredMRMS}a.
\label{fig_brightness_example}}
\end{figure*}
The high spatial resolution of the reflectance data allows for detailed textural information to be encoded. We will encode that texture in $x_2$, but for $x_1$ we wanted to remove as much textural information as possible. To do this, a moving average implemented via a $9 \times 9$ convolution kernel--a $9 \times 9$ matrix whose entries are each $1/81$--was applied to each pixel in the visible imagery. When centered on a pixel, the 80 surrounding reflectance values and the central value are all averaged, and this average is used as the output value for the central pixel location. This process is carried out for all pixels, resulting in a blurred scene of the same size. For this application, the blurred data were then resized using nearest neighbor interpolation from their original resolution, 0.5 km, down to a coarser resolution, 2 km, to match the resolution of the subsequent features. The result was a feature that captured overall brightness in larger regions. To differentiate this information from the reflectance data, $x_1$ will be referred to as the ``brightness'' feature. An example of the brightness feature is displayed in Fig.\ \ref{fig_brightness_example}. This is the blurred and resized version of the reflectance data in Fig.\ \ref{VisibleInfraredMRMS}a.
\subsubsection{Cool Contrast Tiles Feature}
\label{sec:ContrastTiles}
Advancements in the spatial and temporal resolution of data produced by state-of-the-art satellite systems, such as the GOES system, allow us to use visible imagery to (1) identify texture then (2) inform the identification of OTs. Our next feature, $x_2$, was derived by extracting texture information from high-resolution visible imagery. We used the mathematical framework of Gray-Level Co-occurrence Matrices (GLCMs) to quantify texture.

GLCMs capture the spatial relationship between pixel intensities in an image \citep{haralick}. A GLCM is an $N \times N$ matrix where each entry $p(i,j)$ represents the number of times that a pixel with gray level $i$ is adjacent in a fixed direction to and distance from a pixel with gray level $j$. The number $N$ is the number of gray levels in the image, either determined dynamically or set as a hyperparameter. Adjacency is defined by distance and direction--horizontal, vertical, or left/right diagonal--with each distance and direction yielding a GLCM.  Because the matrix entries in a GLCM are dependent on number of pixels in an image, GLCMs are typically normalized so that the sum of its entries is equal to 1. In a normalized GLCM, the entry $p(i,j)$ is the joint probability occurrence of pixel pairs with gray level $i$ and gray level $j$. 

Once a GLCM has been calculated, various statistical measures can be derived which assign a single number to the GLCM matrix. The statistics emphasize different texture characteristics such as orderliness, uniformity, or contrast. Following the work of \cite{Moen2024Gray}, we determined the contrast statistic to be most relevant to our work because strong contrast values were associated with OTs in the images. The contrast statistic sums the matrix entries weighted by the squared difference between gray levels, emphasizing off-diagonal entries and assigning a weight of 0 to diagonal entries. Higher contrast values indicate greater dissimilarity between adjacent pixels.

For each image in our dataset, we split the image into overlapping $8 \times 8$ tiles (using a $4$-pixel stride). This tile size highlights finer-scale texture like OTs, while the stride ensures that texture on tile edges is included. For each tile, we computed GLCMs with adjacency defined by distance one (i.e., touching pixels) in all directions resulting in 8 neighboring pixels for every interior pixel. The images were represented with $N$=256 gray levels, so each GLCM was a $256 \times 256$ matrix regardless of how many gray levels were present in each tile. We then averaged the normalized (distance-one) GLCMs over the four directions and then computed the contrast feature, resulting in a single number for every $4 \times 4$ pixel region (which become our output tiles). As one contrast statistic was calculated for every $4 \times 4$ tile, the image resolution decreased by a factor of four from $0.5 km$ to $2$ km.

Formally, we define the contrast statistic as:
\begin{equation}
        f_{\mathrm{contrast}} = \sum_{i, j = 0}^{N-1}p(i,j)(i-j)^2
         \label{eq: contrast}
\end{equation}
In this statistic, the entries of the GLCM are weighted by $(i-j)^2$, which is larger the further $i$ and $j$ are from each other. Thus, when pixels (in this application, reflectance values) are adjacent which have very different pixel values, $f_{\mathrm{contrast}}$ takes on a larger value.

These contrast values have a lower bound, 0, but not an upper bound. On these data, we found the contrast values to be right-skewed. To solve these issues, we took the log of each contrast value plus one then normalized them to the 0-1 range based on the largest contrast value found in the training data. The same transformation and scaling factor were applied to the validation and test sets, and any resulting values above 1 were truncated. Such large values are rare, so little, if any, information was lost from this truncation.

A potential drawback of utilizing these tiles is the time required to compute them. Though we are confident faster implementations could be derived, ours was quite time-intensive. For the $256 \times 256$ scenes used in this application, processed on a computing resource with 32 Intel(R) Xeon(R) Gold 6134 CPUs running at 3.20GHz and 540 GB of RAM, this feature took roughly 4 seconds to derive, meaning it would take roughly 30 minutes for GOES full-disc ($5,424\times5,424$) imagery.
\begin{figure*}%[htp]
\centering
\noindent\includegraphics[width=0.50\textwidth,angle=0]{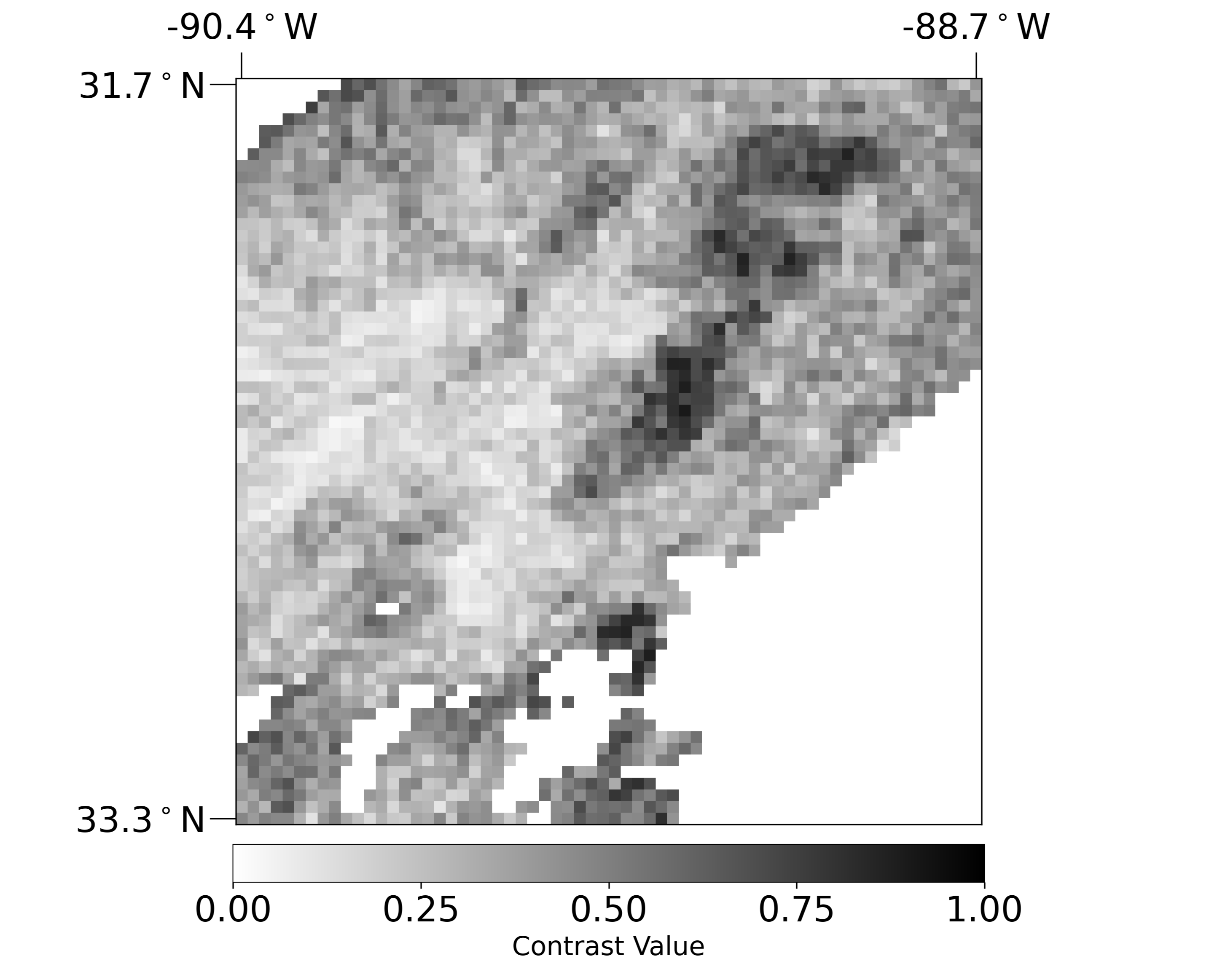}
%\appendcaption{5}
\caption{Cool contrast tiles feature example. Contrast tiles calculated using VIS imagery then made ``cool'' using IR imagery (for this figure, each taken 5 June 2024 at 21:45:00Z). Darker values, those close to one, represent regions with the greatest amount of texture. Lighter values, those close to zero, represent regions with the least amount of texture. Values of exactly zero represent spatial regions corresponding to temperatures greater than 250 K.
\label{fig_GLCM_progression}}
\end{figure*}

The contrast tiles were used to generate the second feature, $x_2$, the ``cool contrast tiles'' feature, seen in Fig.\ \ref{fig_GLCM_progression}. To create $x_2$, contrast tiles that corresponded to brightness temperature values at or below 250 K (i.e., tiles that were relatively ``cool'') were retained and all other tiles (those that corresponded to ``warm'' temperatures) were set to 0.
This modification acts as a high-cloud mask. In the regions where there are no high clouds (i.e., regions of either low clouds or ground), the relationship between texture and OTs is already known and is thus not of interest: we do not expect to find an OT because the region is too warm. The threshold enforces this by removing such scenes from consideration. Similarly to \cite{Lee2021SimplifiedConvection}, we chose a generous (quite warm) brightness temperature at which to threshold. Though OTs are not expected at such warm temperatures, such a low threshold allows for more flexibility during the model alteration process.
\subsubsection{Infrared Feature}
\begin{figure*}%[htp]
\centering
\noindent\includegraphics[width=0.5\textwidth,angle=0]{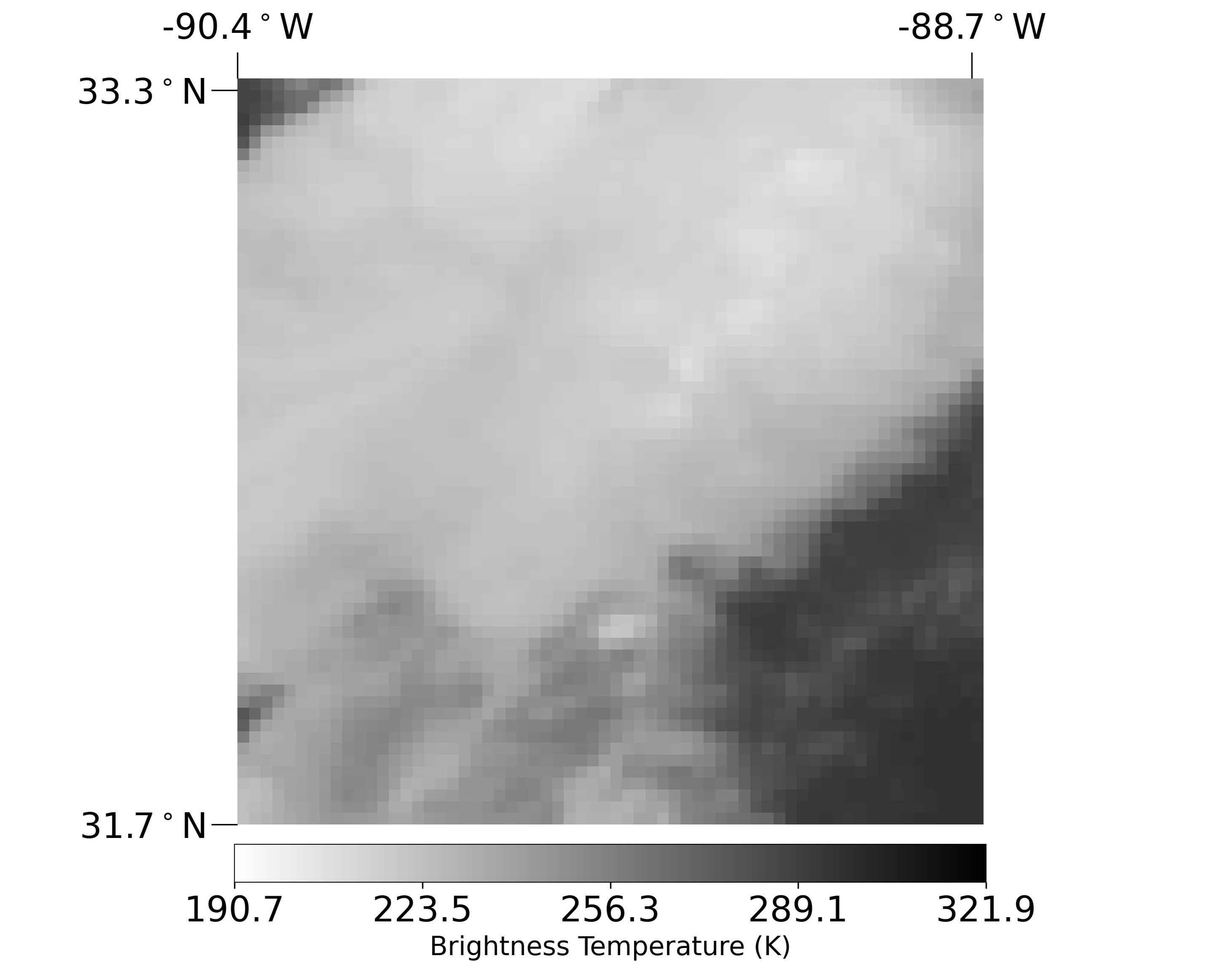}
%\appendcaption{6}
\caption{IR imagery (for this figure, taken 5 June 2024 at 21:45:00Z). This is the same IR imagery displayed in Fig.\ \ref{VisibleInfraredMRMS}b.
\label{fig_infrared_example}}
\end{figure*}

The final feature, $x_3$, was unmodified data from ABI channel 13, the ``clean'' IR longwave window band. Infrared imagery provides information related to surface and cloud-top temperature by measuring the intensity of emitted radiation and converting it into a temperature. In our work, for regions where clouds are present, we use this as a proxy for cloud top height, with higher clouds generally having lower cloud top temperatures and thus lower brightness temperatures. A scene of the infrared feature can be seen in Fig.\ \ref{fig_infrared_example}.
\subsection{Training labels: Multi-Radar Multi-Sensor Data as Convection Labels}
\label{sec:mrms}
To train an ML model to identify OTs, target labels that depict their locations are desirable. One way to achieve this is by identifying echo tops, the highest areas at which precipitation can be detected above the tropopause, and using them as a stand-in for OTs. These echo tops, however, do not always result in accurate OT outlines \citep{Cooney2025AutomatedDetection}. To get a more accurate sense of OT location, human analysts should be brought in and asked to label the scene by hand. \cite{Cooney2025AutomatedDetection}, who chose this route, describe it to be very strenuous.
\begin{figure*}%[htp]
\centering
\noindent\includegraphics[width=0.5\textwidth,angle=0]{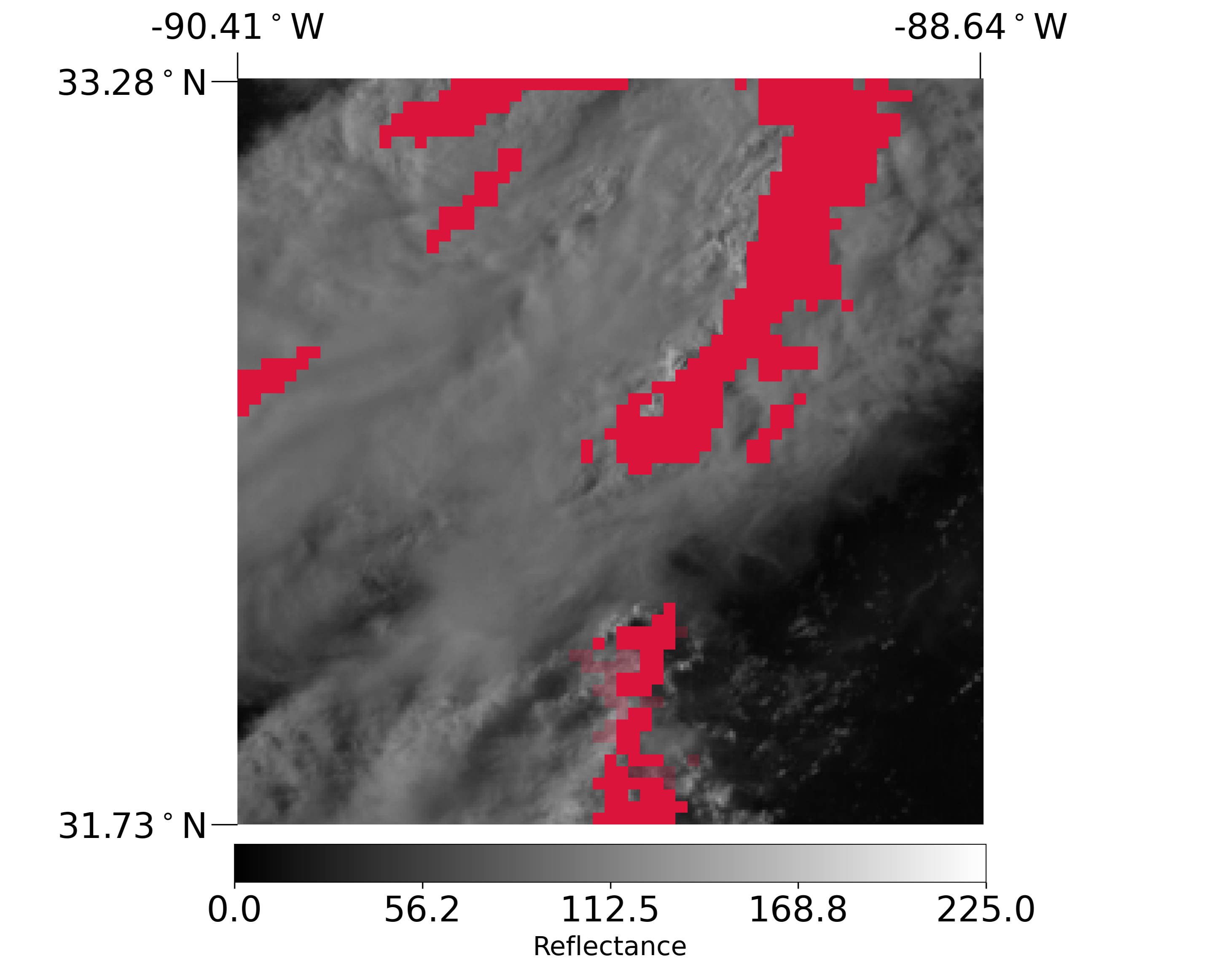}
%\appendcaption{7}
\caption{Labels (in red) corresponding to the location of convection as determined by the MRMS system overlaid on the corresponding reflectance imagery, each captured 5 June 2024 at 21:45:00Z. Full opacity red pixels represent convection occurring at or below 250 K. Translucent red pixels represent the convective labels above 250 K that were removed from consideration. These are the downsampled version of the MRMS labels as displayed in Fig.\ \ref{VisibleInfraredMRMS}c.\label{fig_MRMS_ex}}
\end{figure*}
As an alternative approach, we used labels that were available to us in-house but were not perfect for the task at hand, namely convection labels provided by the MRMS system, created following choices made by \cite{Lee2021SimplifiedConvection}. 

Developed by NOAA National Severe Storms Laboratory, the MRMS system integrates data from ground-based radars, surface and upper air observations, lightning detection system, satellite observations, and forecast models to generate high-resolution meteorological fields. It is designed to support real-time decision-making for severe weather monitoring, aviation operation, and hydrological forecasting. The system provides various fields such as radar reflectivity, precipitation rates, and hydrometeor classification at 1 km spatial resolution and two minute temporal resolution. In this study, the “PrecipFlag” product was used. This product categorizes surface precipitation type into seven categories: warm stratiform rain, snow, convection, hail, cool stratiform rain, tropical stratiform rain, and tropical convective rain. Among these categories, grid points labeled as convection, hail, and tropical convective rain were collectively defined as ``convective'' grid points.

Once rendered and parallax corrected, we reduced the resolution of the labels from 1 km to 2 km using nearest neighbor interpolation. We then imposed a brightness temperature threshold. The brightness temperature (BT) corresponding to each convective pixel was determined using our infrared imagery. If the BT was found to be above 250 K, the convective label was removed from consideration (i.e., set to 0). Only convective pixels corresponding to BTs at or below 250 K were retained. This was done to remove small, isolated regions of convection unlikely to correspond to OTs. In the training set, this removed 30,241 pixels of convection of 377,878 total convective pixels. This removed 11,865 of 197,563 in the validation set and 19,202 of 232,357 in the test set.

Fig.\ \ref{fig_MRMS_ex} exemplifies this process by showing both the full and reduced convection labels atop the corresponding reflectance imagery. Red tiles at full opacity represent convection at or below 250 K while translucent red tiles represent the convective labels above 250 K that were removed from consideration and plotted here for illustration purposes only. For this scene, the only convection that was removed from consideration occurred near the bottom of the scene in the middle.

\subsection{Identifying Convection vs.\ Identifying Overshooting Tops}
\label{sec:TL}
Removing isolated cells of convection was, in part, an attempt to make the \emph{convection} labels better resemble \emph{OT} location labels--though, comparing the large swaths of labeled convection seen in Fig.\ \ref{fig_MRMS_ex} to the isolated OT locations seen in Fig.\ \ref{VisibleInfraredMRMS}(d), we note that this attempt was unsuccessful. The comparison itself, however, sheds light on our decision to use convection labels as training labels for our model. Convection is a necessary precursor for an OT, meaning, if there is an OT, it will be labeled in our dataset as convection.

This becomes useful when we consider that we can edit an EBM's strategies. Because OT location is embedded into the convection labels, even if an EBM is trained to detect the location of \emph{convection} (meaning its strategies are geared toward convection detection), it can be modified to instead detect \emph{OTs} (meaning its strategies are geared toward OT detection). This process can be thought of as another way to accomplish the overall objective of transfer learning \citep{hosna2022transfer}, i.e., the process of developing a model for one task (source task) and then transferring that model to a similar task (target task) through limited modifications. Instead of finding training labels for our task, we aimed to shift from identifying convection to identifying OTs during model development using the following three modifications:
\begin{enumerate}
\item 
    {\bf Feature engineering for OTs:} 
    Selecting only features that focus on identifying OTs, in particular cloud texture, and processing them to emphasize properties that are unique to OTs;
\item 
    {\bf Modification of training labels:} 
    Applying the 250 K temperature threshold to alter the convection labels;
\item
    {\bf Modification of trained model:} 
    Altering the learned strategies of the trained EBM.
\end{enumerate}
Note, however, that as the MRMS data is for convection, it is not immediately suitable to assess accuracy of our algorithm for OTs. See the discussion of this topic in Section \ref{sec:results}.
\section{Model Development}
\label{sec:methods}
We approached the identification of OTs as a pixel-wise binary classification problem. When applied to image data, this can be interpreted as a form of image segmentation. As outlined in Fig.\ \ref{workflows}b the development of an EBM consists of first training a model then visualizing and modifying its strategies as needed based on feedback from domain scientists. Both knowledge of the underlying processes and daily observations of satellite imagery were instrumental in our pursuit to develop an algorithm guided by human knowledge. 

The inclusion of three features--the brightness, cool contrast tiles, and infrared features--necessarily resulted in three primary feature functions and three possible pairwise interaction feature functions, all of which were automatically included.

For reference, using our computing resource (outlined in Section \ref{sec:overshooting_tops_intro}\ref{sec:input_features}), training the EBM took just over 20 minutes on pre-processed data. The memory requirement for the trained EBM was on the order of megabytes, so memory requirements for training and inference are dominated by the size of the data, rather than the model. At inference time, the EBM can be run on the full GOES domain ($5,424\times5,424$ pixels) in just under six seconds.
\subsection{Visualizing EBM Strategies---Plotting Feature Functions and Feature Importance}
\label{sec:feature_functions}
\begin{figure*}%[htp]
\centering
\noindent\includegraphics[width=0.95\textwidth,angle=0]{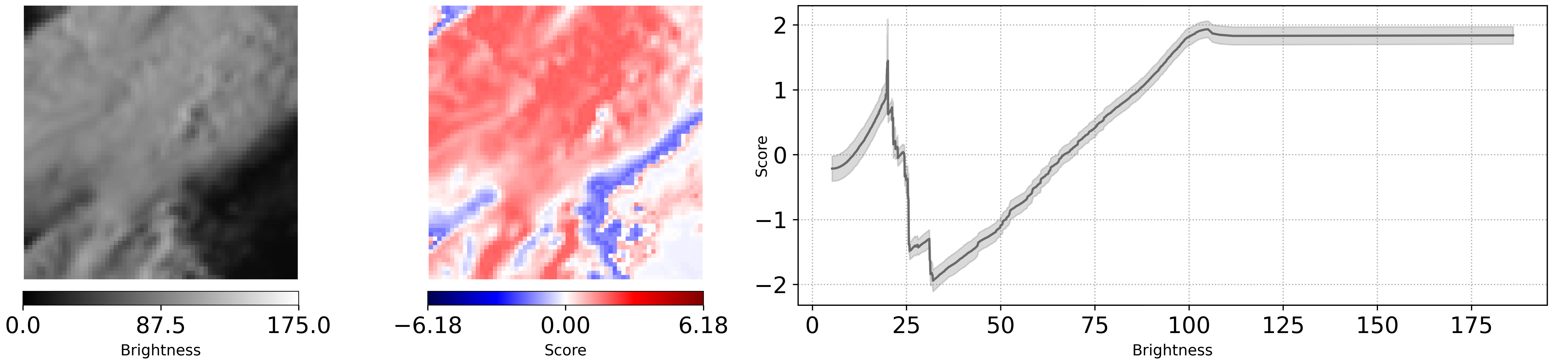}
\\
(a) \hspace*{3.2cm} (b) \hspace*{5.0cm} (c) \hspace*{2cm} 
%\appendcaption{1}
\caption{(a) Brightness feature, (b) feature importance, and (c) corresponding feature function. The feature function, in dark gray, maps brightness values (keys) to scores (values). Shaded error bars representing a measure of uncertainty surround the feature function.
\label{fig_b_i_ff}}
\end{figure*}

After the model is trained, we visualize and analyze model strategies. To illustrate this process, we start by examining the brightness feature function (Fig.\ \ref{fig_b_i_ff}c). The actual strategy relating brightness and convection is the dark line in Fig.\ \ref{fig_b_i_ff}c while the semi-transparent borders represent an automatically generated estimation of model uncertainty from how the EBM is trained \citet{Caruana2015IntelligibleHeatlCare}.

From the lookup table analogy, the $x$-axis values are the keys and the $y$-axis values are the scores. Here, each key relates a range of brightness values to one score. Intuitively, the brightness feature function quantifies the correlation between any given brightness value and the presence of convection. Positive scores indicate a positive relationship and negative scores a negative one while the magnitude indicates the strength.

In Fig.\ \ref{fig_b_i_ff}a we show a brightness feature scene from the test set and its corresponding ``feature importance'' (Fig.\ \ref{fig_b_i_ff}b). The feature importance indicates at each pixel whether brightness alone suggested convection or not. Feature importance is generated by plotting the score value of each pixel from a given feature into a grid; any resemblance to the original feature is not guaranteed.

Interpreting feature importance is straightforward: areas in red represent brightness values positively associated with the presence of convection and those in blue the opposite. The darker the value of the shade, the stronger the relationship.

Viewing a scene of the brightness feature, its feature importance, and the brightness feature function together paints a clear picture: low brightness values are associated with no convection and high brightness values are associated with convection. There is an obvious discrepancy with this reasoning, which we discuss below.
\subsection{Model Alteration--Practical Considerations}
After strategy visualization comes strategy modification. As a precursor, we discuss a potential consequence of being able to edit an EBM: whether or not doing so will lead to overfitting. It is easy to imagine a scenario where strategies altered to address specific problems seen in specific samples leads to poor performance in unseen data. In our experimentation, we did find that when fine-tuned to too high a degree, model performance suffers.

To address this issue, we offer a few practical guidelines. First, we recommend making preliminary alterations to the model based on performance on the training data and only checking model performance on the validation set after an initial round of changes have been made. This allows the validation set to still serve as ``unseen'' data and makes it more likely that the final model will generalize.

Second, we recommend keeping the alterations as general as possible, i.e., not making major changes based on any one sample. After an alteration has been made, we recommend observing performance on a wide variety of samples (particularly ``best'' and ``worst'' cases) in addition to standard performance metrics. Together, these two guidelines help to safeguard against model overfitting and to ensure the model is able to generalize to unseen data.

Finally, when possible, we recommend consulting domain scientists when altering a model. Tying model alterations to physical relationships, however abstract, helps to keep the alterations as general as possible.
\subsection{Model Alteration---Primary Feature Functions}
\label{sec:FF_alteration}
We demonstrate the process of editing a primary feature function on the brightness feature. It is important to note that, as the scene examined is from the test set, it was not used to inform the changes we discuss--it is merely used to exemplify them.

\begin{figure*}%[htp]
\centering
\noindent\includegraphics[width=0.9\textwidth,angle=0]{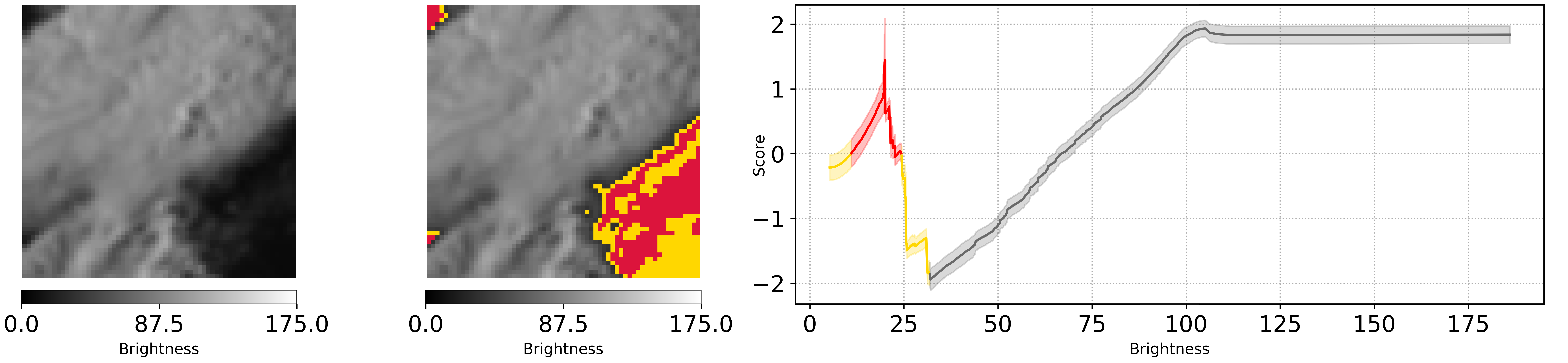}
\\
(a) \hspace*{3.2cm} (b) \hspace*{5.0cm} (c) \hspace*{2cm} 
%\appendcaption{1}
\caption{(a) Brightness, (b) brightness with certain brightness values highlighted in red and yellow, and (c) corresponding  feature function with certain values of brightness highlighted in red and yellow. Brightness values within the feature function's spike that are associated with positive scores are highlighted in red in both (b) and (c) while those associated with negative scores are highlighted in yellow in both (b) and (c).
\label{fig_b_i_h}}
\end{figure*}

The discrepancy we mentioned above refers to the spike in score values seen at low brightness levels in Fig.\ \ref{fig_b_i_ff}c. We initially hypothesized that this was due to the EBM relating shadows to the presence of OTs, but further examination showed it actually corresponds to regions of low-level, small clouds and to darker regions. To see this, Fig.\ \ref{fig_b_i_h}c shows the spike highlighted in two different colors: one for negative score values (yellow) and one for positive score values (red). The corresponding brightness values are highlighted in the same color in Fig.\ \ref{fig_b_i_h}b. Fig.\ \ref{fig_b_i_h}a acts as a reference.

\begin{figure*}%[htp]
\centering
\noindent\includegraphics[width=0.9\textwidth,angle=0]{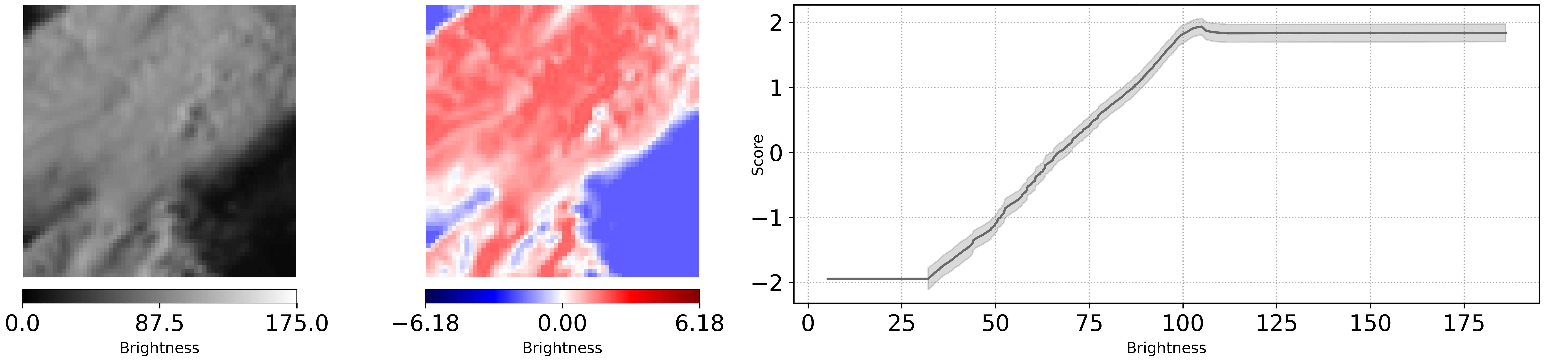}
\\
(a) \hspace*{3.2cm} (b) \hspace*{5.0cm} (c) \hspace*{2cm} 
%\appendcaption{2}
\caption{(a) Brightness, (b) feature importance, and (c) corresponding  feature function mapping pixel values from (a) to pixel values in (b) with shaded error bars to represent a measure of uncertainty. Error bars have been removed from regions of the feature function that were altered. \label{fig_brightness_importance_edited}}
\end{figure*}

Extensive further examination of scenes within the training set revealed that brightness values associated with the spike were never associated with brightness values one might see within an anvil cloud. As such, the decision was made to flatten out the spike to match the lowest score value seen. This alteration and the impact it had on the map of feature importance can be seen in Fig.\ \ref{fig_brightness_importance_edited}.
Fig.\ \ref{fig_brightness_importance_edited} shows the same three plots as Fig.\ \ref{fig_b_i_ff}. In Fig.\ \ref{fig_brightness_importance_edited}c, however, the error bars have been removed from where the feature function received alterations.

\begin{figure*}%[htp]
\centering
\noindent\includegraphics[width=1.0\textwidth,angle=0]{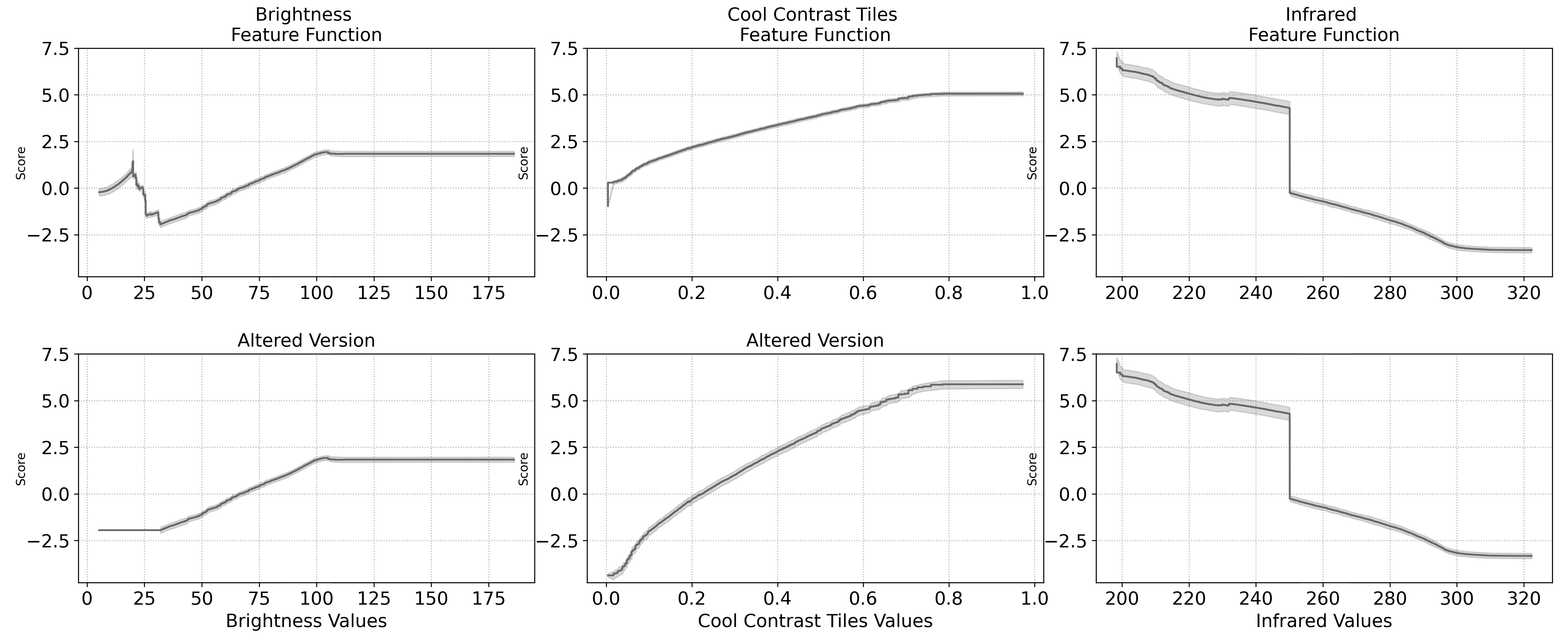}
%\appendcaption{7}
\\
(a) \hspace*{4.5cm} (b) \hspace*{4.5cm} (c)
\caption{Unedited and edited feature functions for all main effect features, (a) brightness, (b) cool contrast tiles, and (c) infrared---unedited functions in the first row and edited functions in the second. Only feature functions (a) and (b) were altered.
\label{AllFFEdited}}
\end{figure*}

\begin{figure*}%[htp]
\centering
\noindent\includegraphics[width=0.9\textwidth,angle=0]{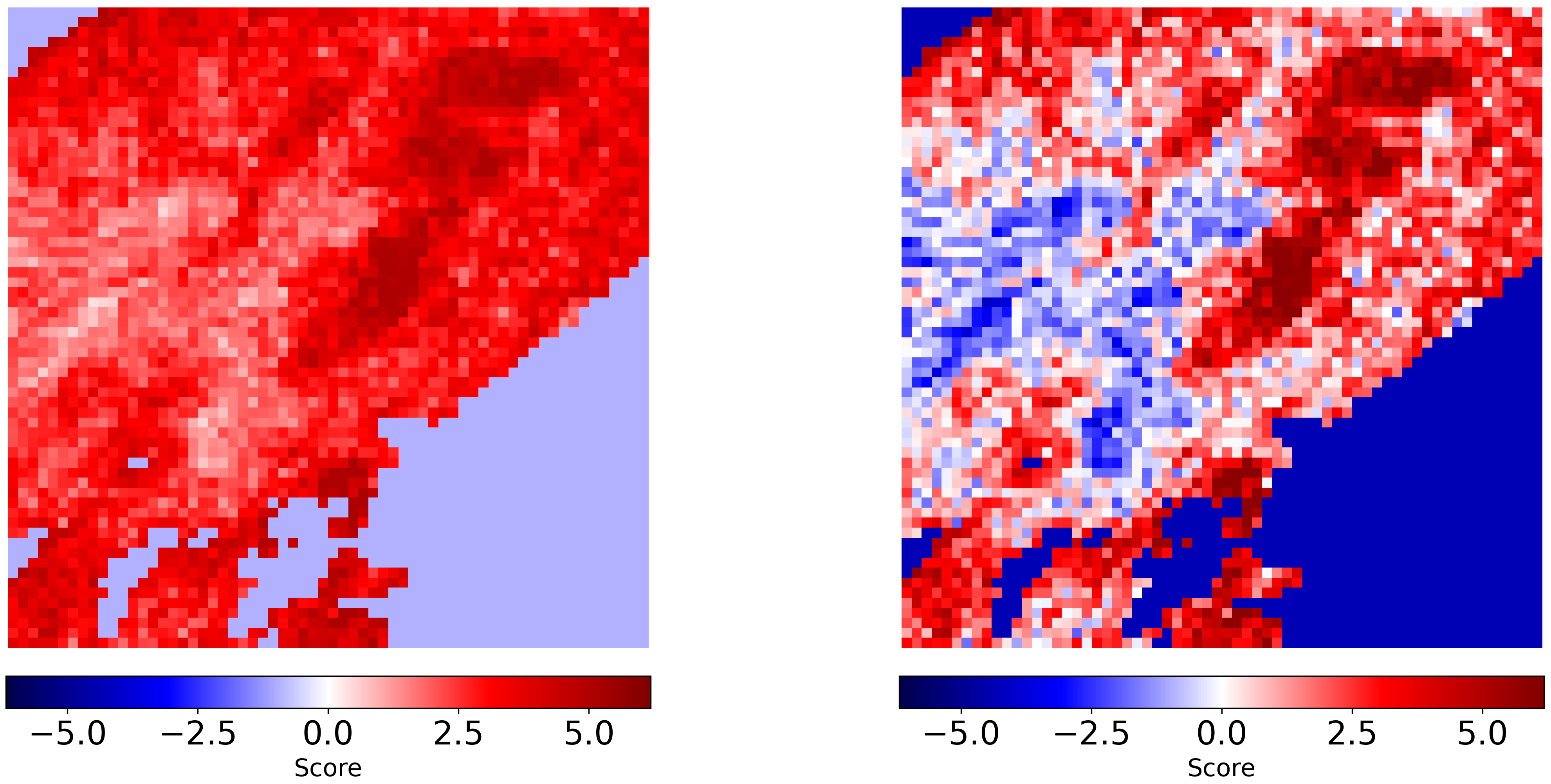}
\\
(a) \hspace*{7.7cm} (b)
\caption{ Map of feature importance for the Cool Contrast Tiles (a) before the model was altered and (b) after the model was altered.
\label{CCT_FI}}
\end{figure*}

Next, we consider all three primary feature functions. The first row of Fig.\ \ref{AllFFEdited} displays the original primary feature functions and the second row the edited primary feature functions.
Of the three, we altered two---the brightness feature function, as described above, and the cool contrast tiles feature function. The cool contrast tiles feature function was scaled to give more ``importance'' (potentially larger score values) to the texture within each scene and then shifted downward to ensure some smaller texture values received small, and even negative, score values. The effect the alteration had can most easily be seen by viewing a map of feature importance, which is shown in Fig.\ \ref{CCT_FI}a.

While the original feature function assigned positive score values to every non-zero texture value, the altered function takes a more nuanced approach. The alterations allowed for small texture values to be associated with negative scores and increased the score given to the largest texture values, which ultimately led to a higher number of overall predictions.

\subsection{Visualizing EBM Strategies--Plotting Interation Feature Functions}
\begin{figure*}%[htp]
\centering
\noindent\includegraphics[width=0.95\textwidth,angle=0]{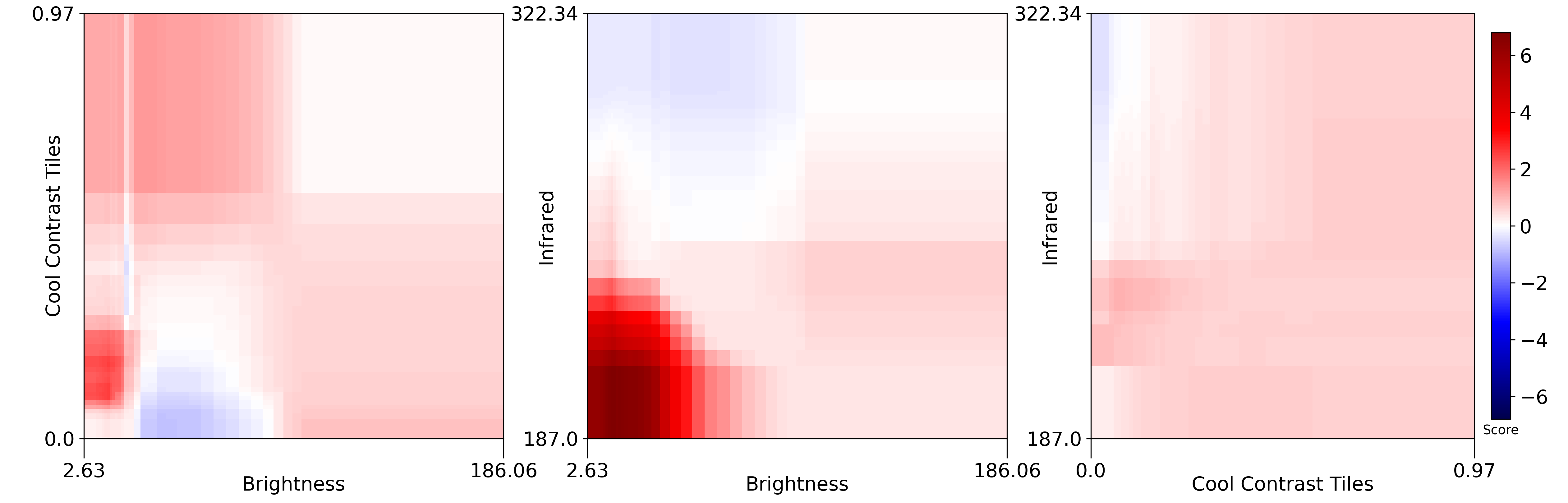}
\\
(a) \hspace*{4.25cm} (b) \hspace*{4.25cm} (c)
\caption{ Feature functions for the three included interactions between (a) the brightness and cool contrast tiles features, (b) the brightness and infrared features, and (c) the cool contrast tiles and infrared features.
\label{InteractionFFs}}
\end{figure*}

Of the three possible pairwise interactions, all three were automatically included in the model. These were between the following pairs of features: brightness and cool contrast tiles, brightness and infrared, and cool contrast tiles and infrared. Each pairwise interaction feature function is represented as a heat map, where each ($x$,$y$) pair key gives the score value for that combination of feature values. Fig.\ \ref{InteractionFFs} displays the three interaction feature functions in the order outlined above. None of the three interaction feature functions were altered.

As outlined, the interaction feature functions are trained in the residuals and \textit{after} the primary functions have been trained. Essentially, the role of the interaction feature functions is to serve as a correction factor between the simpler, single-effect model and the data. The result is that the strategies seen in these feature functions may not as physical as the strategies seen in the primary feature functions. 

\begin{figure*}%[htp]
\centering
\noindent\includegraphics[width=0.95\textwidth,angle=0]{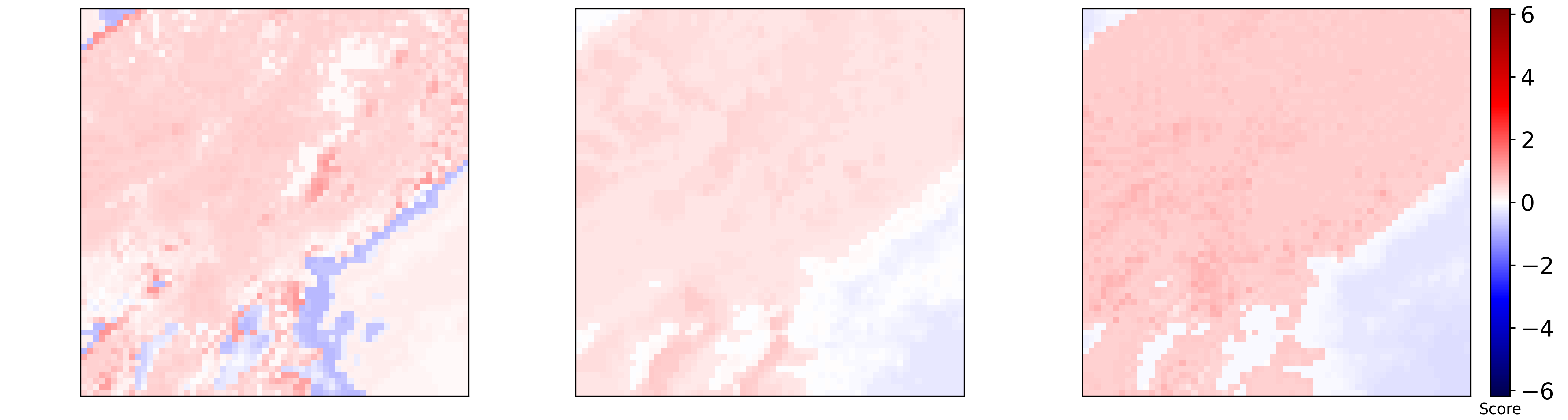}
\\
(a) \hspace*{4.25cm} (b) \hspace*{4.25cm} (c)
\caption{ Maps of feature importance for the interaction feature functions for the interactions between (a) the brightness and cool contrast tiles features, (b) the brightness and infrared features, and (c) the cool contrast tiles and infrared features.
\label{int_importance}}
\end{figure*}

Despite this, some interesting information can be extracted by examining the maps of feature importance corresponding to each interaction feature function seen in Fig.\ \ref{int_importance}. Maps of feature importance for pairwise interactions are similar to the maps of feature importance for the primary features, except that each feature importance map corresponds to two input features. 

In particular, the first (Fig.\ \ref{InteractionFFs}a) pairwise interaction has attributes that, while seemingly counterintuitive, make the model more robust and help to aid OT detection in areas where the primary features cannot.

When examining the interaction between the brightness and cool contrast tiles features, it can be seen that the upper-left corner is red, corresponding to positive scores. This implies that regions of low brightness and high texture are more associated with convection than the primary feature functions would suggest. We confirmed, using processes similar to those described in Section \ref{sec:methods}\ref{sec:FF_alteration}, that these regions correspond to shadows. By comparing the visible imagery seen in Fig.\ \ref{VisibleInfraredMRMS}a and the map of feature importance corresponding to this interaction feature function seen in Fig.\ \ref{int_importance}a, the regions that appear the darkest red near the middle of the scene in Fig.\ \ref{int_importance}a correspond to the regions where OTs cast a shadow. It is important to note, however, that this interaction alone cannot differentiate between shadows cast by an OT and any other dark region featured on the cloud top.

In the interaction between the brightness and infrared features (Fig.\ \ref{InteractionFFs}b), a striking section of red can be seen in the lower-left corner of the feature function. This implies that areas of low brightness and cold temperatures are receiving high, positive scores and indicate the presence of convection. This section, too, can be attributed to helping to detect OT location in shadowed regions, as can be seen in the second case study presented in Section \ref{sec:results}.

The interaction between the Cool Contrast Tiles feature and the Infrared feature (Fig.\ \ref{InteractionFFs}c) does not provide any interesting insights. There is a negative association with convection for nearly all values except tiles that have both little texture and are warm. We believe this unexpected relationship is a correction to the main effects and can be attributed to the fact that the interaction feature functions are trained in the residuals.
\section{Results}
\label{sec:results}
This section focuses on the performance of our final (altered) EBM algorithm. The discussion is split into two parts---the first covers overall model performance metrics (and their limitations) and the second covers three case studies taken from the test set. We acknowledge that these results are not necessarily compelling enough to justify adopting this particular model over existing OT detection algorithms; rather, we wish to use these results as a way to demonstrate how an altered EBM can be evaluated against both quantitative data and human knowledge.
\subsection{Part One: Overall Model Results}
Standard performance-based metrics using existing labels are typically an adequate way to assess how well a model performs on a given task. As discussed in Section \ref{sec:overshooting_tops_intro}\ref{sec:mrms}, however, our methodology is based on labels that mark convection rather than OT locations. Thus, performance metrics using these labels illustrate how well the model is able to detect convection---a task it is not meant for---rather than how well the model is able to detect OTs. 

This causes two issues. First, we expect a high number of misses as a perfect OT-detecting model, by design, should not detect every label denoted as convection by MRMS (see Case I below). Second, we expect a high number of false alarms due to spatial misalignment between OT location and MRMS convection labels (see Case II below).

Furthermore, OTs are being identified on a fairly fine grid of 2 km, meaning some spatial mismatch is inevitable and not necessarily indicative of poor performance. Because of these limitations, metrics are presented for transparency but we do not view them as the best way to judge model performance.

With 2,619 scenes used for testing, there are 10,727,424 pixels. When an identification is made by the model, it is classified as one of the following: a ``hit'' if both the model and the convection labels indicated the presence of convection, a ``correct rejection'' if both the model and the convection labels indicated a lack of convection, a ``false alarm'' if the model indicated the presence of convection but the convection labels did not, and a ``miss'' if the model indicated a lack of convection but the convection labels did not. 
Given the mismatch between the task of detecting OTs and the evaluation of detecting convection, if a scene has 100 pixels labeled as convection by MRMS, but only 30 of those represent OTs, then a perfect OT algorithm would evaluate as having only 30 ``hits'' and 70 ``misses''.

The unedited model achieved 14,959 hits, 10,500,378 correct rejections, 13,891 false alarms, and 198,196 misses. The finalized model achieved 30,755 hits, 10,481,845 correct rejections, 32,424 false alarms, and 182,400 misses. These statistics correspond to a recall--the number of hits over the sum of hits and misses--of 0.070 and 0.144 for the unaltered and altered model, respectively. Furthermore, the model achieved a precision--the number of hits over the sum of hits and false alarms--of 0.519 and 0.487 for the unaltered and altered model, respectively.
The Heidke Skill Score of our model, which compares model performance to a random forecast, increased from 0.119 for the unaltered model to 0.216 for the altered model.
\subsection{Part Two: Case Studies}
\label{sec:case_studies}
We consider three cases and discuss them under the framework of OT identification. The first two cases presented represent scenes where we believe the model performed well at detecting the OTs, while the final case represents a scene where we believe the model performed poorly.  The three cases were selected to illustrate interesting strategies of the final model, including its primary failure modes.
\subsubsection{Case I}
\label{sec:caseI}
\begin{figure*}%[htp]
\centering
\noindent\includegraphics[width=1\textwidth,angle=0]{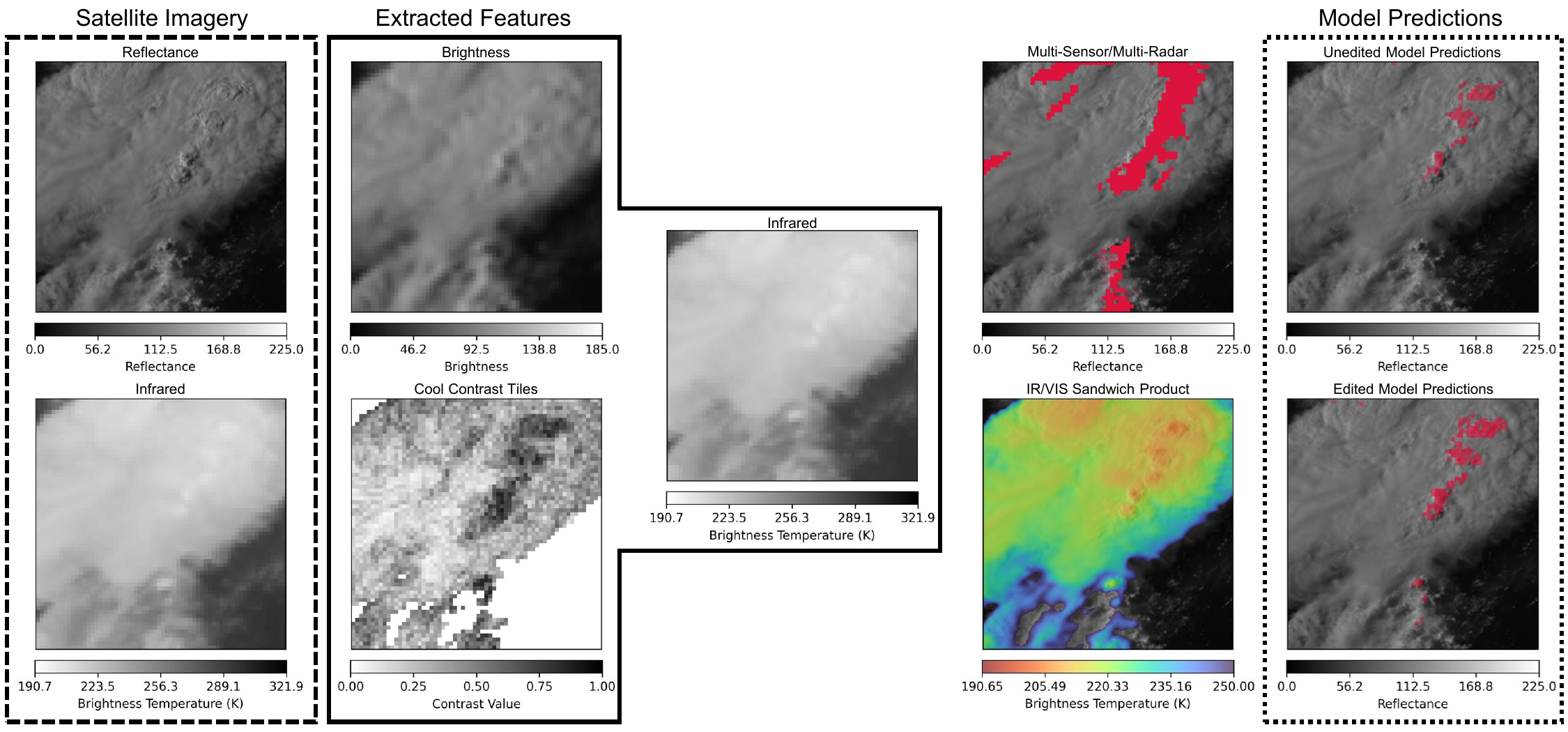}
\\
(a) \hspace*{3.9cm} (b) \hspace*{4.7cm} (c) \hspace*{2.5cm} (d)
%\appendcaption{5}
\caption{EBM workflow stages for selected scene from the test dataset (taken 21:45Z on 5 June 2024) including (a) satellite imagery (reflectance, infrared), (b) extracted features (brightness, cool contrast tiles, infrared), (c) MRMS labels overlaid on visible imagery (top) and IR/VIS Sandwich reference (bottom), and (d) model OT detections (from unedited (top) and edited (bottom) models) overlaid on visible imagery.
\label{fig_766_full}}
\end{figure*}

The first case comes from imagery taken on 5 June 2024 at 21:45:00Z, and is the case we have used repeatedly throughout the paper. Imagery from throughout this discussion has been combined and displayed in Fig.\ \ref{fig_766_full}. Fig.\ \ref{fig_766_full}a displays the satellite imagery used to derive the three features seen in Fig.\ \ref{fig_766_full}b.
The MRMS-derived convection labels can be seen in the first row of Fig.\ \ref{fig_766_full}c. IR/VIS sandwich product imagery is displayed in the second row of Fig.\ \ref{fig_766_full}c. This product is neither used for model development nor quantitative validation, and is used only to provide additional intuition about the scene. Regions where the unedited model detected OTs can be seen in the first row of Fig.\ \ref{fig_766_full}d and regions where the edited model detected OTs can be seen in the second.

Reflectance data shown in the first row of Fig.\ \ref{fig_766_full}a shows five OTs in the upper-right corner. The IR/VIS sandwich product reflects this. Given the strong signal, the unedited model was able to accurately detect their presence and identify their location. Alterations made to the model increased the confidence in these detections and also allowed for three potential OTs to be identified in the lower-middle portion of the scene. The signal these potential OTs provide is much weaker and evidence that supports their status as OTs is not as strong in the IR/VIS sandwich product. Additionally, the model made two isolated erroneous detections in the upper-left corner of the scene where texture is minimal but temperatures are very cold.
\subsubsection{Case II}
\label{sec:caseII}
\begin{figure*}%[htp]
\centering
\noindent\includegraphics[width=1\textwidth,angle=0]{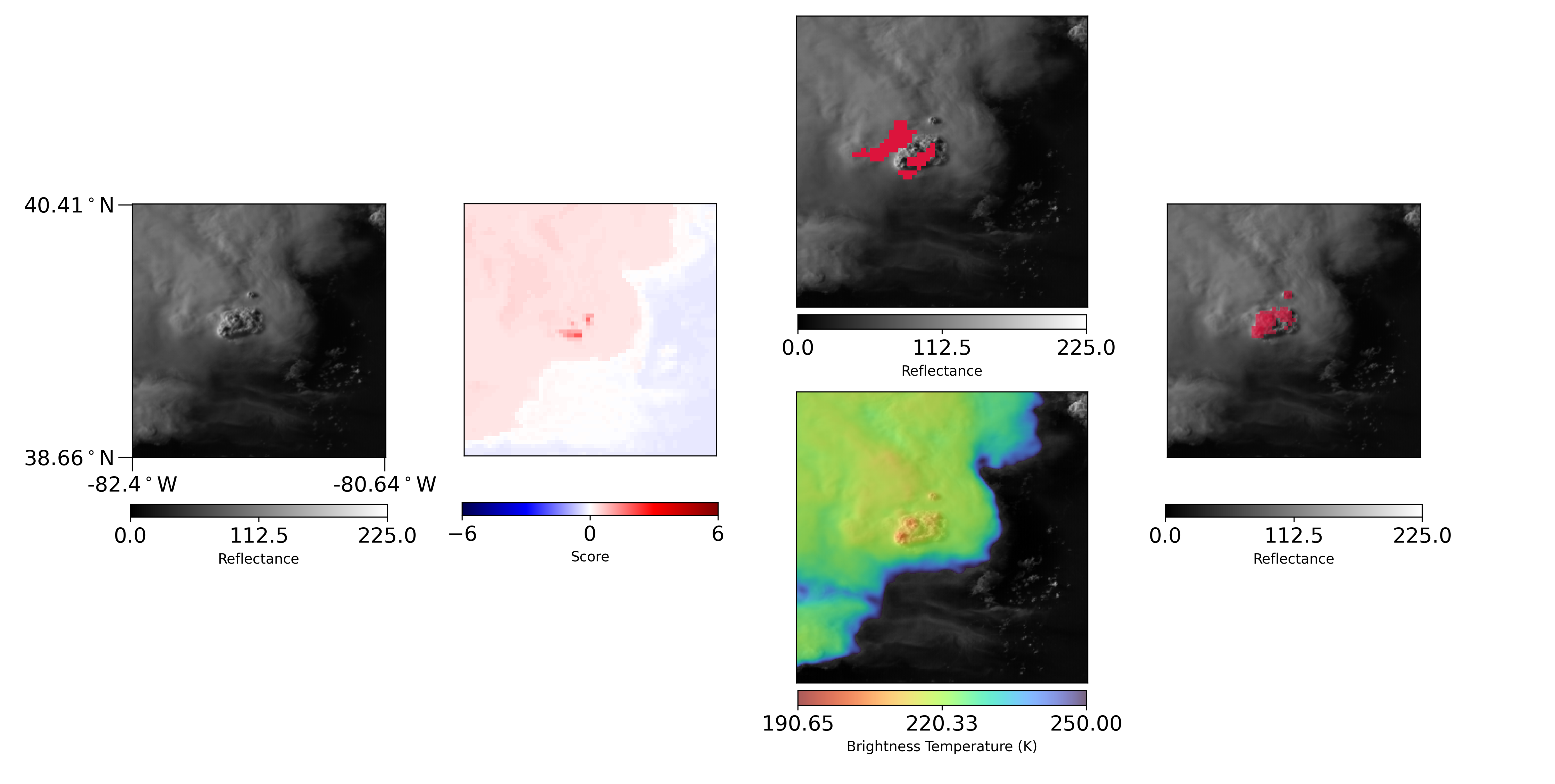}
\\
(a) \hspace*{3cm} (b) \hspace*{3cm} (c) \hspace*{3cm} (d)
%\appendcaption{5}
\caption{Imagery taken 17 June 2024 at 22:30Z including (a) visible imagery, (b) a map of feature importance corresponding to the interaction between the brightness and infrared features, (c) MRMS labels overlaid on visible imagery (top) and IR/VIS sandwich product (bottom), and (d) OT identifications made by the edited model overlaid on visible imagery.
\label{fig_1042}}
\end{figure*}

The second case comes from imagery taken on 17 June 2024 at 22:30:00Z. Fig.\ \ref{fig_1042} displays the corresponding (a) reflectance, (b) map of feature importance corresponding to the interaction between the brightness and infrared features, (c) MRMS labels (top) and IR/VIS sandwich product (bottom), and (d) OT identifications made by the edited model. 
In Fig.\ \ref{fig_1042}a and c, the OTs can be seen almost directly in the center of the scene. They have a bubbly texture and are noticeably colder than their surroundings, while the surrounding anvil is relatively flat and warm. As such, the EBM has no issue detecting their locations.

In this imagery, the direction of the sun causes much of the OT to be covered in shadow. Though the brightness feature function seems to cause dark values to not be associated with OTs, the EBM successfully identifies OT locations. This is, in part, because of the interaction between the brightness and infrared features as seen in Fig.\ \ref{InteractionFFs}b. The associated feature function assigns large, positive scores to pixels that are both dark and cold, i.e., areas of shadow. This effect can be seen in the map of feature importance shown in Fig.\ \ref{fig_1042}b as the areas of shadow have been given large, positive scores.

We also see that, due to the mismatch of the MRMS labels, most of the EBM's detections appear to be false alarms when evaluated against MRMS. Similarly, the MRMS labels resulted in numerous apparent misses. Such issues were seen across the dataset, which negatively impacted the overall model resulted presented.

\subsubsection{Case III}
\begin{figure*}%[htp]
\centering
\noindent\includegraphics[width=1\textwidth,angle=0]{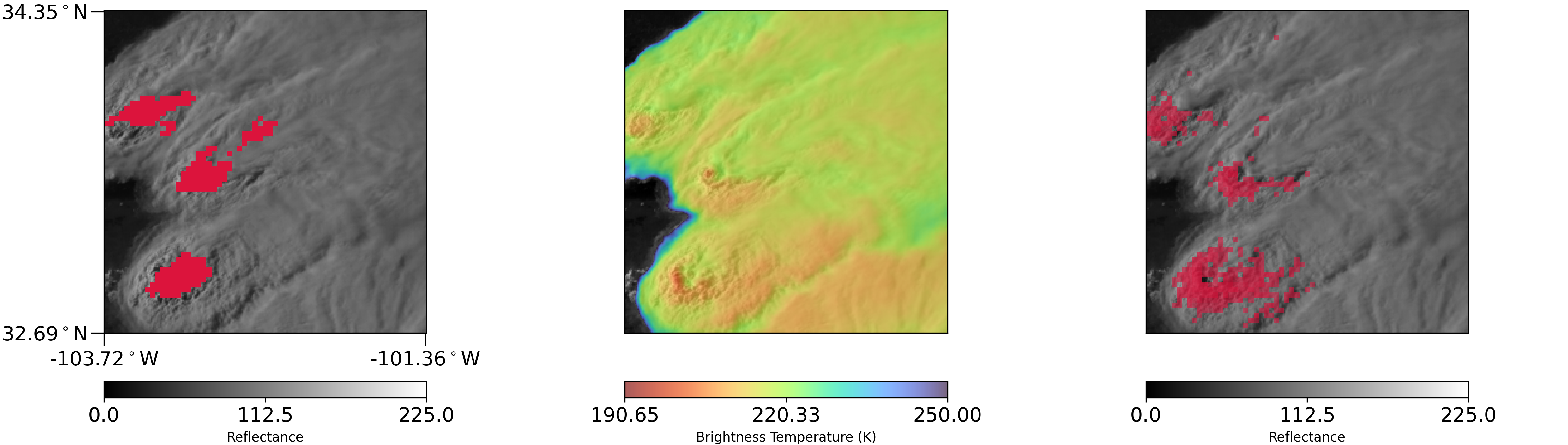}
\\
(a) \hspace*{4.75cm} (b) \hspace*{4.75cm} (c)
%\appendcaption{5}
\caption{Imagery taken at 23:15Z on 30 May 2024 including (a) MRMS labels overlaid on visible imagery, (b) IR/VIS sandwich product, and (c) OT identifications made by the edited model overlaid on visible imagery.
\label{fig_637}}
\end{figure*}

The third case comes from imagery taken on 30 May 2024 at 23:15:00Z. Fig.\ \ref{fig_637} displays the corresponding (a) MRMS labels, (b) IR/VIS sandwich product, and (c) OT identifications made by the edited model. This scene displays three potential OTs as seen in the IR/VIS sandwich product imagery. Surrounding each OT are features consistent with cold U/V shapes and above-anvil cirrus plumes (AACP) \citep{Setvak2010UV}. Much like OTs, these features have been associated with severe weather \citep{Adler1985Severe}.

Our goal, however, was to detect OTs specifically and not features associated with severe weather in general. This case exemplifies how the EBM is unable to distinguish between OTs and these other features--a common failure mode of the model. This failure mode is due to limitations in the chosen modeling framework as, without any additionally features, the EBM is physically unable to differentiate OTs and cold U/Vs as both are cold, textured, and, with ample daylight, bright. Had a model capable of learning more complex relationships been used, this may not have been an issue--even with the same three features. Because we are able to view all of the model's learned relationships, however, we were {\it able} to determine that this failure mode is due to the modeling framework and not a faulty relationship.

We use this case to exemplify another failure mode of the model regarding the limitations of how we defined texture. By nature, the contrast statistic returns large values where there is high contrast. The bubbly texture we seek for identifying OTs features high contrast, but high contrast values are not limited to such texture. Near the middle of this scene, the EBM predicted an OT just to the right of its actual location (note that it did, also, encapsulate the OT). This area to the right does not feature bubbly texture, but there {\it is} high contrast--i.e., bright values directly next to dark ones--resulting from the AACP. 

In this scene, the cold U/V shape provided cool temperatures near the AACP. In other scenes with high contrast not due to OTs, the model fails when the cloud-top is much colder than average. Because of limitations in the modeling framework, the EBM cannot differentiate cloud tops that are bubbly and cold because there is an OT and cloud tops that feature high contrast and are cold because they are higher up in the troposphere. For example, in another scene, the EBM predicted an OT where there was an anvil merger because the anvils were, on average, colder than the model was used to seeing.

Neither issue can be fixed for this model because the strategy being used is correct. Any attempt to lessen the importance of the IR or Cool Contrast Tiles feature results in fewer OTs being detected. Additional features could be of use--specifically tropopause height--but the model considered here was purposefully kept simple to enhance interpretability. Including such features is an opportunity for future work.
\section{Conclusions and Future Work}
\label{sec:conclusions}
In this work, we make the case for the use of EBMs in combination with physics-informed feature engineering to yield interpretable ML algorithms for certain meteorological applications. This approach has several advantages, including, but not limited to
(1) the ability to fully understand the strategies used by the ML algorithm when making identifications, exposing potential failure modes;
(2) the opportunity to adjust its strategies to more closely match the strategies expected based on domain knowledge; and
(3) the ability to develop a generalizable model from just a few data samples, or, if data for a similar task is available, utilizing those instead in a way analogous to transfer learning.

We have illustrated how these advantageous aspects of the EBM framework can aid in the approach of detecting OT locations from satellite imagery. We emphasize, however, that this application of EBMs was only possible due to feature engineering that first simplified the task at hand. Nevertheless, we believe that this method has the potential to be used in a wide variety of meteorological applications.

At first sight, the identification and tuning of EBM model strategies, which is the part of the EBM development process illustrated in Section \ref{sec:methods}, may appear to be a lot of extra work, especially when compared to the hands-off training procedure of a comparable neural network model. One should keep in mind, however, that for a neural network model, the identification of strategies should come as a separate step {\it after} its training is completed, e.g., using XAI methods, but that step is often neglected, since it is nearly impossible to detect most of its strategies anyway.  EBMs should thus not be dismissed for enabling and, in fact, requiring this important step during their development process. In other words, this step is simply the price to pay to obtain an interpretable model. 

For the application of identifying OTs, the next step in this research should be the creation of a large hand-labeled data set that identifies OTs in GOES visible imagery.  Creating such a labeled data set is a larger effort, but it is needed to fully evaluate how well the EBM model matches human labeling. More generally, much work remains to further explore the use of EBMs in the field of meteorology in terms of both identifying the most suitable applications and developing a larger range of engineered features---endeavors we hope will serve to further improve the performance of these models.

\clearpage
%%%%%%%%%%%%%%%%%%%%%%%%%%%%%%%%%%%%%%%%%%%%%%%%%%%%%%%%%%%%%%%%%%%%%
% ACKNOWLEDGMENTS
%%%%%%%%%%%%%%%%%%%%%%%%%%%%%%%%%%%%%%%%%%%%%%%%%%%%%%%%%%%%%%%%%%%%%
\acknowledgments
This material is based upon work supported by the National Science Foundation under AI Institute Grant No.\ 2019758 and CAIG grant No.\ 2425923; and by the Machine Learning Strategic Initiative at the Cooperative Institute for Research in the Atmosphere at Colorado State University.

%%%%%%%%%%%%%%%%%%%%%%%%%%%%%%%%%%%%%%%%%%%%%%%%%%%%%%%%%%%%%%%%%%%%%
% DATA AVAILABILITY STATEMENT
%%%%%%%%%%%%%%%%%%%%%%%%%%%%%%%%%%%%%%%%%%%%%%%%%%%%%%%%%%%%%%%%%%%%%
% 
%
\datastatement
The dataset and python code used to train, validate, and test the EBM model will be made publicly available before publication.

%%%%%%%%%%%%%%%%%%%%%%%%%%%%%%%%%%%%%%%%%%%%%%%%%%%%%%%%%%%%%%%%%%%%%
% APPENDIXES
%%%%%%%%%%%%%%%%%%%%%%%%%%%%%%%%%%%%%%%%%%%%%%%%%%%%%%%%%%%%%%%%%%%%%
%
%% If only one appendix, use

%\appendix

%% If more than one appendix, use \appendix[<letter>], e.g.,

%\appendix[A] 

%% Appendix title is necessary! For appendix title:

%\appendixtitle{Title of Appendix}

%%% Appendix section numbering (note, skip \section and begin with \subsection)
%
% \subsection{First primary heading}

% \subsubsection{First secondary heading}

% \paragraph{First tertiary heading}

%%%%%%%%%%%%%%%%%%%%%%%%%%%%%%%%%%%%%%%%%%%%%%%%%%%%%%%%%%%%%%%%%%%%%
% REFERENCES
%%%%%%%%%%%%%%%%%%%%%%%%%%%%%%%%%%%%%%%%%%%%%%%%%%%%%%%%%%%%%%%%%%%%%
% Make your BibTeX bibliography by using these commands:
\bibliographystyle{ametsocV6}
\bibliography{references}

\end{document}